%% file: main.tex
\pdfoutput=1
\documentclass[10pt,twocolumn,letterpaper]{article}

\usepackage[pagenumbers]{preprint}

\usepackage{multirow}
\usepackage{colortbl}
\usepackage{tabularx}
\usepackage{amsthm}
\newtheorem{proposition}{Proposition}
\definecolor{refgray}{gray}{0.93}
\definecolor{ourblue}{rgb}{0.90,0.94,0.99}
\newcommand{\ours}{VIBRA\xspace}

\definecolor{linkblue}{rgb}{0.21,0.49,0.74}
\usepackage[pagebackref,breaklinks,colorlinks,allcolors=linkblue]{hyperref}

\title{VIBRA: Backward-Compatible Cross-Modal Retrieval with Consensus Barycenters}

\author{Shubing Yang\\
University of Washington\\
{\tt\small sueyoung@uw.edu}
\and
Xutong Liu\\
University of Washington\\
{\tt\small xutongl@uw.edu}
\and
Dongfang Zhao\\
University of Washington\\
{\tt\small dzhao@cs.washington.edu}
}

\begin{document}
\maketitle

\begin{abstract}
Backward-compatible retrieval enables a new encoder to query the gallery that an old model has already encoded, so that a cross-modal retrieval system can be upgraded without re-encoding billions of items. Existing compatible methods either fit the old query and stay near the accuracy of the old model, or run the old and the new encoder for every query. We propose VIBRA, which matches relevance on the old gallery, corrects the errors of a single model with the barycenter of several views, which is itself a query in the old space, and trains the new encoder to inherit this consensus, so that models absent at deployment still shape every query. We show that the barycenter gains margin over the average of its views exactly when their hardest irrelevant items differ, and we characterize the share of this gain that one encoder can reach. On five benchmarks with five upgrades among six CLIP, MetaCLIP and SigLIP2 backbones, VIBRA is the best compatible method in all 50 settings, improving Recall@1 by up to 5.6 points over the strongest baseline and by up to 18.0 points over the old model, while adding less than 1 ms to the latency of the new encoder. Code is available at \url{https://github.com/githubullet/VIBRA}.
\end{abstract}

\section{Introduction}
\label{sec:intro}

\begin{figure}[t]
\centering
\includegraphics[width=\linewidth]{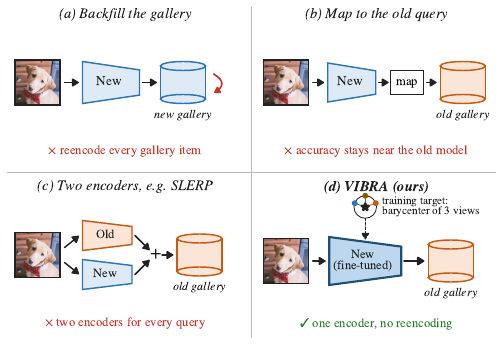}
\caption{Four ways to upgrade the query encoder. Only \ours (d) keeps the gallery and one encoder per query while improving accuracy.}
\label{fig:teaser}
\end{figure}

Backward compatibility allows a cross-modal retrieval system to adopt a new encoder while its billions of gallery items keep the embeddings of the old model.  Vision and language models such as CLIP~\cite{radford2021clip,jia2021align} encode every gallery item once and answer each query by a nearest neighbor search over the stored vectors~\cite{johnson2019faiss}. These models improve quickly through new data, architectures and training objectives~\cite{xu2024metaclip,tschannen2025siglip2}, yet two independently trained models produce embeddings that cannot be compared with each other~\cite{shen2020bct}. Without compatibility, adopting a new model requires re-encoding the whole gallery, known as backfilling (Figure~\ref{fig:teaser}(a)), which costs hundreds of GPU hours for billions of items and is impossible once the raw items are gone~\cite{ricci2026slerp}. In cross-modal search the old gallery even holds the other modality, so a new image encoder must be compared with captions that only the old text encoder has seen.

Existing compatible methods avoid backfilling and can be grouped into three categories (Figure~\ref{fig:teaser}), which train the new model for compatibility, map its queries to the old query, or run both encoders for every query. Backward-compatible training, introduced by BCT~\cite{shen2020bct} and extended by later work~\cite{ramanujan2022fct,zhou2023bt2}, trains the new model with a compatibility loss toward the old one, and query side maps~\cite{vejendla2025drift,maiorca2023lst,ricci2025lorth} keep a pretrained new model frozen and translate its queries into the old space with a light map (Figure~\ref{fig:teaser}(b)). 

XBT~\cite{jang2025xbt} tunes the new encoder through a projection pretrained on text. These methods fit the old query of each input, so they inherit the errors of the old model and their accuracy stays close to it. SLERP~\cite{ricci2026slerp} interpolates between the aligned new query and the old query and gains accuracy, but it runs the old encoder next to the new one (Figure~\ref{fig:teaser}(c)), which raises the latency of every query.

Our key insight is that both limitations stem from the training target. A target defined by the old query reproduces at best the ranking of the old model, including its errors, whereas a target defined by relevance, the gallery items that a query should retrieve, can exceed it. Several models correct one another when they make errors on different items, so one encoder can learn from a teacher that combines the new, the old and an auxiliary encoder.

\begin{figure*}[t]
\centering
\includegraphics[width=\textwidth]{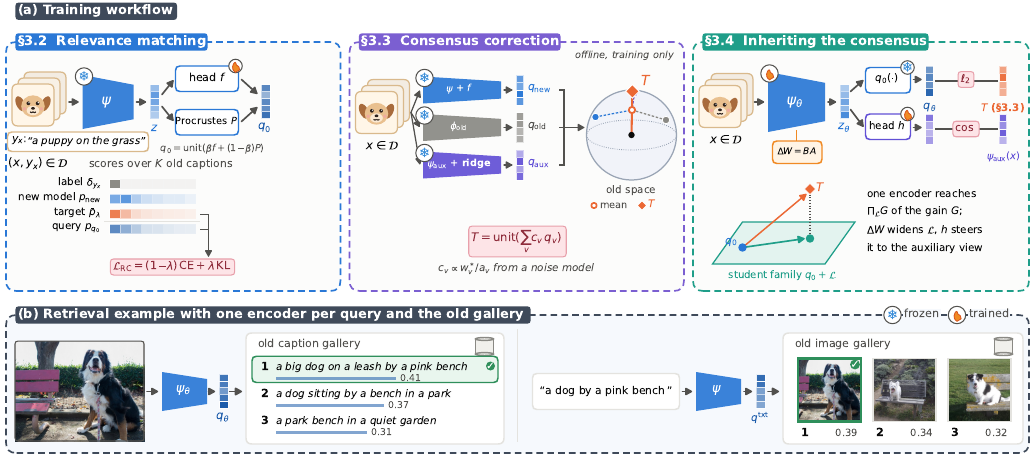}
\caption{Overview of \ours. (a) Training workflow. Sec.~\ref{sec:optimal} fits the stage~1 query $q_0$ with the relevance loss $\mathcal{L}_{\mathrm{RC}}$; Sec.~\ref{sec:consensus} pools the views $q_v$ offline into the teacher $T$ with weights $c_v$; Sec.~\ref{sec:distill} trains a low rank update $\Delta W$ so that $q_\theta$ matches $T$, reaching the share $\Pi_{\mathcal{L}}G$ of the gain $G$. (b) Retrieval example, in which an image query and a text query each run one encoder and retrieve the correct item from the old gallery.}
\label{fig:overview}
\end{figure*}

To address both limitations, we propose \ours, which matches relevance on the old gallery, corrects the errors of a single model with the barycenter of several views, itself a query in the old space, and lets the new encoder inherit this consensus, so that models absent at deployment still shape every query (Figure~\ref{fig:overview}). Relevance matching derives the ideal query from the optimality condition of a listwise retrieval loss and estimates relevance by mixing the paired caption with the ranking of the new model. Consensus correction pools the queries of the new, the old and an auxiliary encoder into a view barycenter, with weights that a noise model of the views fixes before training. Consensus inheritance trains a low rank update of the new encoder, steered toward the auxiliary view, to reproduce the barycenter and merges it into the encoder, so that deployment runs one encoder.

Our analysis shows that the margin of the barycenter exceeds the average margin of its views exactly when their hardest irrelevant items differ, and it gives the share of this gain that one encoder can reach. Pooling therefore corrects errors that the views make on different items, no weighting corrects errors that they share, and one encoder recovers the part of this gain that lies in the directions it can reach.

In experiments on five benchmarks and five upgrades among six CLIP, MetaCLIP and SigLIP2 backbones, \ours ranks first among compatible methods in all 50 settings, with Recall@1 gains of up to 5.6 points over the strongest baseline and up to 18.0 points over the old model, at a latency within 1 ms of the new encoder. On average it raises Recall@1 by 6.5 points over the old model and by 2.3 points over SLERP, which runs two encoders.

Our contributions are as follows.
\begin{itemize}
\item We propose \ours, a single encoder approach to cross-modal compatibility that learns a barycenter of several encoders derived from the retrieval objective, while leaving the old gallery unchanged (Section~\ref{sec:method}).
\item We develop a theory of the compatible target, which is defined by relevance on the old gallery (Section~\ref{sec:optimal}), gains a dispersion margin over its views exactly when their hardest irrelevant items differ (Proposition~\ref{prop:margin}), and is inherited by one encoder up to a projection of this gain (Section~\ref{sec:distill}).
\item We evaluate \ours on five benchmarks and five upgrades, where it is the best compatible method in all 50 settings, improves Recall@1 by up to 5.6 points over the strongest baseline and by up to 18.0 points over the old model, and keeps the latency of a single encoder (Section~\ref{sec:exp}).
\end{itemize}

\section{Related Work}
\label{sec:related}

\paragraph{Backward-compatible training.}
Backward-compatible training makes a new model comparable with features that an old model has already computed~\cite{shen2020bct,duggal2021compat}. BCT~\cite{shen2020bct} trains the new embedding model with an influence loss through the classifier of the old model, so that its queries can be compared directly with old gallery features. Later methods improve the balance between compatibility and the accuracy of the new model through class center alignment~\cite{meng2021lce}, adversarial learning~\cite{pan2023advbct}, basis transformation~\cite{zhou2023bt2}, stationary classifiers~\cite{biondi2023cores,biondi2024stationary} or orthogonal transformations~\cite{ricci2024orthogonal}, while others lower the cost of backfilling itself~\cite{ramanujan2022fct,zhang2022hotrefresh,jaeckle2023fastfill,seo2023online,zhang2023darwinian}. These methods address image retrieval with class labels and usually train the new model from scratch, whereas we fine-tune an existing pretrained vision and language model across modalities without re-encoding any gallery item.

\paragraph{Query side maps.}
Query side maps keep the new model frozen and translate its queries into the old space. Drift-Adapter~\cite{vejendla2025drift} fits Procrustes, low rank affine or residual MLP adapters on paired embeddings, $\lambda$-orth~\cite{ricci2025lorth} learns a transformation with relaxed orthogonality, and latent space translation~\cite{maiorca2023lst,moschella2023relative,huh2024platonic} estimates the map from semantically aligned anchors; related adapters address continual text retrievers~\cite{goswami2025qdc} and embeddings of multimodal large language models~\cite{byun2026mkp}. All of them fit the old query of the same input, so their objective contains no mechanism that corrects the errors of the old model (Section~\ref{sec:optimal}), and they can only use information that the output of the new encoder already contains (Section~\ref{sec:distill}).

\paragraph{Cross-modal compatibility.}
Cross-modal compatibility extends backward compatibility to vision and language models, whose gallery holds the other modality. XBT~\cite{jang2025xbt} introduced it with a projection module pretrained on text and a low rank update of the new encoder. TaCA~\cite{zhang2023taca} inserts a compatible adapter into an upgraded visual encoder, and SLERP~\cite{ricci2026slerp} interpolates on the sphere between an aligned new query and the old query, which requires both encoders at query time. \ours keeps the deployment constraint of XBT, one query encoder and an untouched gallery, and also adapts the new encoder with a low rank update, but derives what the update learns from the retrieval objective and from several views.

\paragraph{Soft targets and distillation.}
Soft targets and distillation transfer knowledge from other models into the model being trained. ALBEF~\cite{li2021albef} mixes one hot labels with the similarity distribution of a momentum copy of the model to soften noisy image and text pairs; our stage~1 target has the same form, but its soft ranking comes from another model, the new one, it is matched on the old gallery, and the mixture follows from the optimality condition of the retrieval objective (Section~\ref{sec:optimal}). Beyond matching the outputs or features of a teacher~\cite{hinton2015distilling,romero2015fitnets}, relational and ranking distillation~\cite{park2019rkd,tang2018ranking} transfer the structure among items, and ensembles and several teachers~\cite{krogh1995ensembles,you2017multiteacher} are combined with weights that are usually set by hand; for vision and language models, distillation has mainly served compression~\cite{wu2023tinyclip,yang2024clipkd}. In \ours, the teacher lives in the old space and the student must stay compatible with a frozen gallery; the teacher is a barycenter of views, the form that a single query can reproduce, and its weights follow from the noise of the views, as in the combination of forecasts~\cite{bates1969combination}.

\section{Method}
\label{sec:method}

\subsection{Overview}
\label{sec:overview}

We upgrade the query encoder of a retrieval system whose gallery must stay as it is. The gallery $\mathcal{G}=\{g_y\}$ holds unit vectors $g_y=\phi_{\mathrm{old}}(y)\in\mathbb{R}^{d}$ computed by the old model $\phi_{\mathrm{old}}$ for every item $y$. A new model $\psi$ gives features $z=\psi(x)\in\mathbb{R}^{d'}$ of a query $x$. Both $\phi_{\mathrm{old}}$ and $\psi$ denote the image or text tower that matches their input. We build a query map $q(x)$ on the unit sphere $\mathbb{S}^{d-1}$ that runs one network, initialized from $\psi$, and ranks the gallery by the score $s_q(x,y)=\langle q(x),g_y\rangle$; we write $\mathrm{unit}(v)=v/\|v\|$. Training may run any encoder on a set $\mathcal{D}$ of image and caption pairs, whereas deployment runs only $q$. We describe image queries and treat text queries in Section~\ref{sec:distill}.

\ours has three components, one per subsection (Figure~\ref{fig:overview}). Relevance matching replaces the imitation of the old query (Section~\ref{sec:optimal}), a barycenter of views corrects the errors of a single model (Section~\ref{sec:consensus}), and the new encoder is trained on that barycenter so that one encoder suffices (Section~\ref{sec:distill}).

\subsection{Relevance Matching with Ranking Consistency}
\label{sec:optimal}

The stage~1 query matches relevance on the old gallery through the loss of Eq.~\eqref{eq:rcd} (Figure~\ref{fig:overview}(a), left). Query side maps imitate the old query, and imitation cannot exceed the old model. Procrustes alignment, Drift-Adapter, LST and $\lambda$-orth fit the new query to the old query $\phi_{\mathrm{old}}(x)$ of the same input. A perfect fit gives $q(x)=\phi_{\mathrm{old}}(x)$ and therefore exactly the ranking and the recall of the old model on its own gallery (Old $\to$ old): every item the old model ranks wrongly with confidence stays wrong. An imperfect fit can land somewhat above or below the old model, but the objective itself contains no mechanism that moves the query toward relevant items the old model misses. To exceed the old model, the target must be defined by the \emph{relevance} of the gallery items.

The retrieval objective defines such a target through the relevance of the gallery items. Recall at any rank is piecewise constant in $q$, so we work with its standard listwise surrogate~\cite{cao2007listnet}. Let $p^*(\cdot\mid x)$ be the relevance distribution of query $x$, which spreads unit mass over the gallery items relevant to $x$, and let $p_q(y\mid x)=\exp(s_q(x,y)/\tau)/\sum_{y'}\exp(s_q(x,y')/\tau)$ be the distribution that a query induces over the gallery, with temperature $\tau$. The cross entropy $\mathcal{L}(q;x)=-\sum_y p^*(y\mid x)\log p_q(y\mid x)$ is convex in $q$, and its gradient $\tfrac1\tau\big(\mathbb{E}_{y\sim p_q}[g_y]-\mathbb{E}_{y\sim p^*}[g_y]\big)$ vanishes exactly at a query $q^\star$ with
\begin{equation*}
\mathbb{E}_{y\sim p_{q^\star}(\cdot\mid x)}[g_y]=\mathbb{E}_{y\sim p^*(\cdot\mid x)}[g_y],
\end{equation*}
that is, when the barycenter of the gallery weighted by the query coincides with the barycenter of the relevant items; for unit queries the condition holds for the component tangent to the sphere. The optimal query is thus determined by relevance measured on the old gallery, and we adopt this condition as the target. During training the gallery is a pool of $K$ candidate captions per image encoded by the old model, and every distribution below is taken over this pool.

We estimate the unobserved relevance $p^*$ by mixing two estimates that a training pair offers, whose errors differ. The label $\delta_{y_x}$, which puts all mass on the caption $y_x$ paired with $x$, is correct about that caption but treats every other relevant caption as irrelevant. The ranking of the new model, $p_{\mathrm{new}}(y\mid x)=\mathrm{softmax}_y(\langle\psi(x),\psi(y)\rangle/\tau)$ over the same items, recognizes such captions but carries the errors of the new model. We therefore use the mixture $\bar p_\lambda=(1-\lambda)\,\delta_{y_x}+\lambda\,p_{\mathrm{new}}$ with a weight $\lambda\in[0,1]$. Because the cross entropy is linear in the target distribution, the loss equals, up to a constant independent of $q$,
\begin{equation}
\mathcal{L}_{\mathrm{RC}}(q;x)=(1-\lambda)\,\mathrm{CE}(\delta_{y_x},p_q)+\lambda\,\mathrm{KL}(p_{\mathrm{new}}\,\|\,p_q),
\label{eq:rcd}
\end{equation}
the contrastive loss~\cite{oord2018cpc} on the label plus a ranking consistency term toward the new model, both measured on the old gallery.

We fit this loss with a linear head on the frozen new features and anchor the head with the Procrustes map~\cite{schonemann1966procrustes}. We minimize~\eqref{eq:rcd} with a linear head $f(z)=\mathrm{unit}(Wz+b)$ with weights $W$ and bias $b$ on the frozen new features and use $\lambda=\tfrac12$, the equal weighting of the two terms. At a stationary point the residual of the optimality condition for $\bar p_\lambda$, projected onto the tangent space of the sphere and weighted by $1/\|Wz+b\|$, is uncorrelated with every coordinate of $z$; the head therefore matches the estimated relevance on average, in the least squares sense over linear functions of $z$. The head is fit on finite candidate pools and can distort the geometry of the new space, whereas the orthogonal Procrustes map, the matrix $P\in\mathbb{R}^{d'\times d}$ with orthonormal rows or columns that best maps the new features of the training images onto their old image embeddings, only rotates and projects that space. We therefore average the head with $P$, which serves as an anchor, $q_0(z)=\mathrm{unit}\big(\beta\,f(z)+(1-\beta)\,\mathrm{unit}(P^\top z)\big)$ with a fixed weight $\beta$. The stage~1 query $q_0$ uses only the new encoder, and so it keeps the errors of the new model, which the next section addresses.

\subsection{Consensus Correction with View Barycenters}
\label{sec:consensus}

The teacher is the barycenter of three views, Eq.~\eqref{eq:teacher}, which corrects the errors that the views make on different items (Figure~\ref{fig:overview}(a), middle). Views from other encoders can correct the errors of a single model. Any query estimated from one model ranks wrongly whatever that model ranks wrongly with confidence. Other encoders see the same image differently, and during training they are available. We call a \emph{view} any unit query in the old space: the stage~1 head alone, $q_{\mathrm{new}}=f(z)$, which we index as view $0$, the old query $q_{\mathrm{old}}=\phi_{\mathrm{old}}(x)$, and auxiliary encoders $\psi_{\mathrm{aux}}$ mapped to the old space by ridge regression onto the old image embeddings, $q_{\mathrm{aux}}=\mathrm{unit}(R\,\psi_{\mathrm{aux}}(x)+e)$ with ridge weights $R$ and offset $e$. Using views to remove the errors of one model raises two questions: which targets combine several views and can still be represented by a single query, and which errors of the views such a target can correct.

\begin{figure}[t]
\centering
\includegraphics[width=\linewidth]{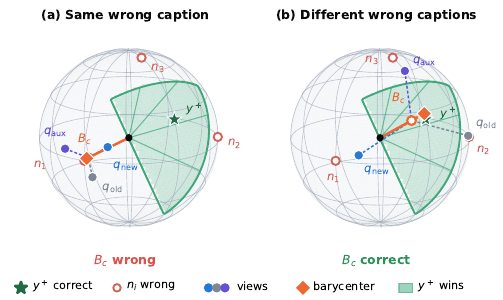}
\caption{Consensus correction on the sphere. $B_c$ projects the mean of the views (hollow point) onto the sphere and ranks $y^+$ first inside the green cone. (a) Views share their hardest irrelevant caption, so $D=0$ and $B_c$ fails. (b) Their hardest irrelevant captions differ, so $D>0$ and $B_c$ succeeds while every view fails.}
\label{fig:sphere}
\end{figure}

Every weighted fusion of view scores is reproduced by a single query, the view barycenter. A query interacts with the gallery only through its scores. Fusing the scores of the views with weights $c$ in the simplex gives, by linearity, $\sum_v c_v\,s_v(x,y)=\langle\sum_v c_v q_v,g_y\rangle$ with $s_v=\langle q_v,g_y\rangle$, so every weighted fusion of view scores is reproduced exactly by a single query. For $\sum_v c_v q_v\ne0$, its normalized form is the \emph{view barycenter}, the point of the sphere with the smallest weighted squared chordal distance to the views,
\begin{equation}
B_c(x)=\operatorname*{arg\,min}_{u\in\mathbb{S}^{d-1}}\sum_v c_v\|u-q_v(x)\|^2=\mathrm{unit}\Big(\sum_v c_v\,q_v(x)\Big).
\label{eq:barycenter}
\end{equation}
Normalization rescales all scores of a query by the same factor, so the barycenter ranks the gallery exactly as the fused scores do. A target that combines several views and is representable by one query therefore takes the form of a view barycenter, one vector per training image. It remains to determine which errors of the views it corrects. Whether the barycenter corrects an error depends on which irrelevant items each view ranks above $y^+$ (Figure~\ref{fig:sphere}).

\begin{proposition}[Barycentric margin]
\label{prop:margin}
Fix a query with a relevant item $y^+$ and irrelevant items $\mathcal{N}$. For each view let $\Delta_v(n)=s_v(y^+)-s_v(n)$ for $n\in\mathcal{N}$, let $\Delta_{v,(k)}$ be its $k$th smallest value and $m_v=\Delta_{v,(1)}$ the margin of the view, and let $\Delta_B=\sum_v c_v\Delta_v$ be the gaps of the barycenter, up to a positive factor, with margin $m_B=\Delta_{B,(1)}$. Then:\\
(i) the margin of the barycenter decomposes as
\begin{equation*}
m_B=\textstyle\sum_v c_v\,m_v+D,\qquad 0\le D\le\sum_{v\ne0}c_v\,\gamma_v(\hat n),
\end{equation*}
where $\gamma_v(n)=\Delta_v(n)-m_v\ge0$ and $\hat n$ is the hardest irrelevant item of view $0$, and $D>0$ if and only if the views with $c_v>0$ do not share their hardest irrelevant item;\\
(ii) for all integers $k_v\ge1$ with $\sum_v(k_v-1)+1=k$, $\Delta_{B,(k)}\ge\sum_v c_v\,\Delta_{v,(k_v)}$, so $y^+$ is among the top $k$ of the barycenter whenever the right side is positive.
\end{proposition}

\begin{proof}[Proof sketch]
(i) For every irrelevant item $n$, $\Delta_B(n)=\sum_v c_v\Delta_v(n)\ge\sum_v c_v m_v$, and taking the minimum over $n$ gives $D\ge0$. Evaluating $\Delta_B$ at $\hat n$, where view $0$ attains its margin, gives the upper bound. If $n^\star$ attains $m_B$, then $D=\sum_v c_v(\Delta_v(n^\star)-m_v)$ is a sum of nonnegative terms, which vanishes exactly when $n^\star$ is the hardest item of every view with $c_v>0$. (ii) View $v$ has at most $k_v-1$ items with $\Delta_v<\Delta_{v,(k_v)}$, so at most $k-1$ items lie below some threshold, and among the $k$ items with the smallest barycentric gaps one lies above all of them. The full proof is in Section~S3 of the supplementary material.
\end{proof}

The proposition identifies the quantity that decides whether pooling corrects errors. Pooling never lowers the average margin of the views, and it adds the \emph{dispersion gain} $D$ exactly when the views do not share their hardest irrelevant item. In that case the barycenter can rank $y^+$ first even when some or all views rank it lower, since a positive $D$ can outweigh negative view margins. Conversely, if every view ranks $y^+$ below one common irrelevant item $n_c$, then $\Delta_B(n_c)=\sum_v c_v\Delta_v(n_c)<0$ for all weights, so no barycenter of these views corrects the error. Part (ii) extends the same statement from the first rank to recall at rank $k$. The proposition thus prescribes the views: they must differ in their hardest irrelevant items, which the three views of the new, the old and an auxiliary model trained on other data provide, and three views suffice, since the bound on $D$ does not grow with the number of views and the influence of a further view decays as $1/(V+1)$ with $V$ views.

A noise model of the views sets the weights of the barycenter and predicts which views help. Proposition~\ref{prop:margin} turns the choice of views into a criterion: a candidate view can correct the new view only on queries where its hardest irrelevant item differs from that of the new view, and since the proposition holds for any weights, the weights only need to reflect the reliability of each view. We model the score of view $v$ as $s_v(x,y)=a_v\,r(x,y)+b_v(x)+\varepsilon_v(x,y)$, where $r\in\{0,1\}$ marks relevance, $a_v>0$ is the separation of the view, $b_v(x)$ is an offset shared by all items of a query, and $\varepsilon_v$ is noise with zero mean. The rescaled noises of the views have the covariance matrix $\Sigma_{uv}=\mathrm{Cov}(\varepsilon_u/a_u,\varepsilon_v/a_v)$ over the irrelevant items, assumed equal for all queries. The fused score $\sum_v c_v s_v$ separates relevant from irrelevant items by $\sum_v c_v a_v$; writing $w_v\propto c_v a_v$ for the weights normalized to this separation, the fusion with the smallest noise variance~\cite{bates1969combination} has
\begin{equation*}
w^*=\frac{\Sigma^{-1}\mathbf{1}}{\mathbf{1}^\top\Sigma^{-1}\mathbf{1}},\qquad c_v\propto w^*_v/a_v,
\end{equation*}
where $\mathbf{1}$ is the vector of ones. We use these weights when all entries of $w^*$ are nonnegative, so that $c$ lies in the simplex, and otherwise the minimizer over nonnegative weights. For view $0$ and one extra view whose rescaled noises have standard deviations $\sigma_0$ and $\sigma_a$ and correlation $\rho$, the relative reduction of noise variance is
\begin{equation*}
\Gamma=1-\frac{\sigma_a^2(1-\rho^2)}{\sigma_0^2+\sigma_a^2-2\rho\sigma_0\sigma_a}.
\end{equation*}
The gain depends on the correlation and the noise ratio together; with nonnegative weights an extra view lowers the noise of view $0$ exactly when $\rho<\sigma_0/\sigma_a$. Both $c$ and $\Gamma$ are estimated from held out pairs before training, so the views and their weights are fixed in advance. We use three views, $q_{\mathrm{new}}$, $q_{\mathrm{old}}$ and LAION CLIP ViT-H/14~\cite{schuhmann2022laion,cherti2023openclip} as $q_{\mathrm{aux}}$, and the teacher
\begin{equation}
T(x)=B_c(x)=\mathrm{unit}\big(c_{\mathrm{new}}\,q_{\mathrm{new}}+c_{\mathrm{old}}\,q_{\mathrm{old}}+c_{\mathrm{aux}}\,q_{\mathrm{aux}}\big).
\label{eq:teacher}
\end{equation}

\subsection{Inheriting the Consensus in a Single Encoder}
\label{sec:distill}

We train the new encoder itself to produce the barycenter (Figure~\ref{fig:overview}(a), right), since the teacher needs the old and the auxiliary encoders, which deployment forbids. A view that is a function of $z$, such as the head $f$ or the Procrustes map, can be pooled at deployment at no cost; a view that needs another encoder cannot. Let $\psi_\theta$ be the new encoder with a low rank update $\Delta W=BA$~\cite{hu2022lora} of the attention query and value projections of the image tower, whose factors form $\theta$, $z_\theta=\psi_\theta(x)$, and $q_\theta=q_0(z_\theta)$ the stage~1 query function, with $f$ and $P$ frozen, applied to the adapted features. The student minimizes
\begin{equation}
\mathbb{E}_{x\sim\mathcal{D}}\Big[\|q_\theta(x)-T(x)\|^2+\mu\big(1-\cos(h(z_\theta),\psi_{\mathrm{aux}}(x))\big)\Big],
\label{eq:full}
\end{equation}
where $\mu$ weights the auxiliary term and $h$ is a linear head trained jointly and discarded after training. The target, the trainable family and the auxiliary term of~\eqref{eq:full} follow from the analysis below.

Matching the barycenter as a vector keeps the ranking of the teacher in every gallery and regresses onto all views at once. If $\|q_\theta(x)-T(x)\|\le\delta$, every score against a unit gallery vector moves by at most $\delta$, so the student keeps the first ranked item of the teacher whenever the teacher margin exceeds $2\delta$, in every gallery, including galleries never seen in training. A loss on rankings constrains only the candidate pools seen in training. Matching the vector also matches all views at once: for every unit query $q$, with $\bar q=\sum_v c_v q_v$,
\begin{equation*}
\|q-T\|^2=\frac{\sum_v c_v\|q-q_v\|^2-\kappa(x)}{\|\bar q(x)\|},
\end{equation*}
where $\kappa=2(1-\|\bar q\|)$ depends only on the query, so the first term of~\eqref{eq:full} is a joint regression onto the views, weighted per query by $1/\|\bar q(x)\|$, which is largest when the views disagree, that is, where Proposition~\ref{prop:margin} locates the dispersion gain.

The first term of~\eqref{eq:full} is also a least squares problem over the student family, which shows what a single encoder can reach and why the encoder itself must be adapted. Writing the teacher as $T=q_0+G$, where $G$ is the gain of the barycenter over the stage~1 query from which the student starts, stacking the outputs of all training images into one vector, and taking the family to first order and without the unit norm constraint as $q_0+\mathcal{L}$, with $\mathcal{L}$ the span of the Jacobian columns of the trainable parameters, the projection theorem gives the minimizer $q_0+\Pi_{\mathcal{L}}G$, with $\Pi_{\mathcal{L}}$ the orthogonal projection onto $\mathcal{L}$, which captures the fraction
\begin{equation*}
\eta=\|\Pi_{\mathcal{L}}G\|^2/\|G\|^2
\end{equation*}
of the gain, the share of $G$ that the family explains. A head on the frozen features $z$ spans only maps of $z$, while the gain comes from views whose errors differ from those of the new model, which such maps express poorly. Adapting the encoder enlarges $\mathcal{L}$ in the directions that matter. The same result implies that recovery is partial whenever $G$ has a component outside $\mathcal{L}$, and that the student can only learn a gain that the barycenter shows on the training images.

The second term of~\eqref{eq:full} steers the adaptation toward the auxiliary view. It asks the adapted features to predict the native embedding of the auxiliary encoder through the head $h$, which is initialized by ridge regression from $z$. It adds no module at deployment, since $h$ is discarded, and reaches the query only through the features, where it favors the directions of $\mathcal{L}$ that encode the auxiliary view, from which part of the gain $G$ comes.

Text queries need no student. Both available views are functions of the new text features $t$: the stage~1 text query $q_0^{\mathrm{txt}}(t)$, built as in Section~\ref{sec:optimal} with old images as the gallery, and a map $M$ trained contrastively on captions of the training corpus, whose target is the old text embedding of the same caption, so that it needs captions only. The text query is their barycenter $q^{\mathrm{txt}}=\mathrm{unit}\big(\alpha\,q_0^{\mathrm{txt}}(t)+(1-\alpha)\,\mathrm{unit}(M(t))\big)$ with a fixed weight $\alpha$; since neither view needs another encoder, it is computed directly at deployment.

Training has three stages: fit the stage~1 heads with~\eqref{eq:rcd} and the text map $M$; estimate the weights $c$ on held out pairs and compute the teacher~\eqref{eq:teacher} offline for every training image; train the auxiliary head and only the low rank factors of the encoder with~\eqref{eq:full}, which keeps the upgrade cheap. At deployment an image query runs the new image encoder with the update merged into its weights, followed by $q_0$, and a text query runs the new text encoder followed by the text barycenter. The cost is one forward pass of the new encoder plus a small map, the same as any single encoder baseline.

\section{Experiments}
\label{sec:exp}

\paragraph{Setup.}
We evaluate five upgrades of a deployed encoder, always against the old gallery. We upgrade the old CLIP ViT-B/32~\cite{radford2021clip} to CLIP ViT-L/14, CLIP ViT-B/16, SigLIP2 SO400M/14~\cite{zhai2023siglip,tschannen2025siglip2} and MetaCLIP ViT-L/14~\cite{xu2024metaclip}, and the old MetaCLIP ViT-B/32 to SigLIP2 SO400M/14. All training uses CC3M~\cite{sharma2018cc3m} only, with 100K images for the student and 2.35M captions for the text map. All experiments run on a single NVIDIA A100 GPU.

\paragraph{Benchmarks.}
We evaluate image to text and text to image retrieval by Recall@1 on Flickr30k~\cite{young2014flickr} and COCO~\cite{lin2014coco} with the large galleries of~\cite{jang2025xbt}, NoCaps~\cite{agrawal2019nocaps}, CxC~\cite{parekh2021cxc} and the English part of XM3600~\cite{thapliyal2022xm3600}.

\paragraph{Baselines.}
We compare with five compatible methods. Drift-Adapter~\cite{vejendla2025drift}, LST~\cite{maiorca2023lst} and $\lambda$-orth~\cite{ricci2025lorth} are query side maps, XBT~\cite{jang2025xbt} fine-tunes the new encoder, and SLERP~\cite{ricci2026slerp} runs the old and the new encoder for every query.

\paragraph{Hyperparameters.}
All hyperparameters are fixed on held out pairs before any test evaluation. We set $K=1024$, $\lambda=\tfrac12$, $\beta=0.75$, $\alpha=0.75$, $\tau=0.03$ and $\mu=1$, take the teacher weights $c\propto w^*/a$ derived from the noise model on held out CC3M pairs, and train LoRA of rank 8 for 2 epochs at learning rate $10^{-4}$.

\subsection{Accuracy}
\label{sec:main_results}
\begin{table*}[t]
\centering
\footnotesize
\setlength{\tabcolsep}{4pt}
\renewcommand{\arraystretch}{0.92}
\caption{Recall@1 (\%) against the old gallery. Gray: references. $^\dagger$ published; $^\ddagger$ our reproduction of XBT. \textbf{Bold}: best compatible method.}
\label{tab:main}
\begin{tabularx}{\textwidth}{l*{10}{>{\raggedleft\arraybackslash}X}}
\toprule
& \multicolumn{5}{c}{Image $\to$ text} & \multicolumn{5}{c}{Text $\to$ image} \\
\cmidrule(lr){2-6} \cmidrule(lr){7-11}
Method & Flickr30k & COCO & NoCaps & CxC & XM3600 & Flickr30k & COCO & NoCaps & CxC & XM3600 \\
\midrule
\multicolumn{11}{l}{\textit{CLIP ViT-B/32 $\to$ CLIP ViT-L/14}} \\
\rowcolor{refgray} Old $\to$ old & 40.7 & 28.6 & 71.2 & 51.6 & 53.3 & 21.8 & 14.5 & 45.3 & 32.3 & 38.7 \\
\rowcolor{refgray} New $\to$ new & 48.7 & 34.4 & 73.4 & 58.0 & 54.1 & 28.2 & 18.7 & 47.8 & 38.2 & 42.5 \\
Drift-Adapter~\cite{vejendla2025drift} & 42.4 & 29.1 & 67.4 & 52.4 & 45.9 & 21.7 & 14.1 & 43.9 & 31.9 & 38.6 \\
XBT~\cite{jang2025xbt} & 42.5$^\dagger$ & 30.7$^\dagger$ & 75.0$^\dagger$ & 50.8$^\ddagger$ & 50.8$^\ddagger$ & 22.4$^\dagger$ & 15.6$^\dagger$ & 48.0$^\dagger$ & 33.9$^\ddagger$ & 41.6$^\ddagger$ \\
SLERP~\cite{ricci2026slerp} & 48.0$^\dagger$ & 33.4$^\dagger$ & 74.0$^\dagger$ & 56.0 & 55.6 & 23.3$^\dagger$ & 15.3$^\dagger$ & 46.6$^\dagger$ & 33.8 & 40.5 \\
LST~\cite{maiorca2023lst} & 40.4 & 28.6 & 67.3 & 52.4 & 45.9 & 20.9 & 13.7 & 42.9 & 30.9 & 37.1 \\
$\lambda$-orth~\cite{ricci2025lorth} & 35.6 & 25.7 & 64.0 & 49.5 & 47.2 & 21.2 & 13.8 & 44.4 & 31.4 & 40.3 \\
\rowcolor{ourblue} \ours{} (ours) & \textbf{51.6} & \textbf{36.0} & \textbf{76.1} & \textbf{59.6} & \textbf{56.1} & \textbf{25.9} & \textbf{17.3} & \textbf{50.8} & \textbf{36.8} & \textbf{45.1} \\
\midrule
\multicolumn{11}{l}{\textit{CLIP ViT-B/32 $\to$ SigLIP2 SO400M/14}} \\
\rowcolor{refgray} Old $\to$ old & 40.7 & 28.6 & 71.2 & 51.6 & 53.3 & 21.8 & 14.5 & 45.3 & 32.3 & 38.7 \\
\rowcolor{refgray} New $\to$ new & 69.1 & 50.4 & 88.5 & 72.4 & 65.8 & 51.5 & 35.1 & 70.1 & 56.8 & 58.8 \\
Drift-Adapter~\cite{vejendla2025drift} & 50.6 & 34.3 & 74.2 & 58.2 & 52.9 & 24.2 & 15.7 & 46.9 & 34.7 & 39.9 \\
XBT~\cite{jang2025xbt} & 38.5$^\ddagger$ & 27.9$^\ddagger$ & 71.6$^\ddagger$ & 51.8$^\ddagger$ & 54.2$^\ddagger$ & 23.2$^\ddagger$ & 16.2$^\ddagger$ & 49.7$^\ddagger$ & 35.7$^\ddagger$ & 43.4$^\ddagger$ \\
SLERP~\cite{ricci2026slerp} & 51.0$^\dagger$ & 37.4$^\dagger$ & 78.6$^\dagger$ & 61.7 & 58.9 & 25.8$^\dagger$ & 17.2$^\dagger$ & 50.6$^\dagger$ & 36.5 & 43.2 \\
LST~\cite{maiorca2023lst} & 46.1 & 31.5 & 71.6 & 55.3 & 49.9 & 22.9 & 15.0 & 44.9 & 33.4 & 38.4 \\
$\lambda$-orth~\cite{ricci2025lorth} & 42.2 & 30.4 & 70.6 & 55.4 & 53.8 & 25.5 & 16.5 & 50.2 & 35.7 & 43.6 \\
\rowcolor{ourblue} \ours{} (ours) & \textbf{54.0} & \textbf{38.5} & \textbf{79.5} & \textbf{62.2} & \textbf{60.8} & \textbf{30.3} & \textbf{20.3} & \textbf{56.2} & \textbf{41.2} & \textbf{48.7} \\
\midrule
\multicolumn{11}{l}{\textit{MetaCLIP ViT-B/32 $\to$ SigLIP2 SO400M/14}} \\
\rowcolor{refgray} Old $\to$ old & 43.3 & 33.0 & 77.9 & 56.7 & 59.2 & 28.0 & 19.3 & 55.9 & 40.1 & 49.9 \\
\rowcolor{refgray} New $\to$ new & 69.1 & 50.4 & 88.5 & 72.4 & 65.8 & 51.5 & 35.1 & 70.1 & 56.8 & 58.8 \\
Drift-Adapter~\cite{vejendla2025drift} & 58.5 & 40.9 & 81.0 & 64.5 & 58.8 & 26.8 & 18.2 & 53.3 & 38.6 & 45.9 \\
XBT~\cite{jang2025xbt} & 49.0$^\ddagger$ & 34.4$^\ddagger$ & 79.8$^\ddagger$ & 59.0$^\ddagger$ & 57.8$^\ddagger$ & 22.7$^\ddagger$ & 17.0$^\ddagger$ & 53.1$^\ddagger$ & 37.6$^\ddagger$ & 45.9$^\ddagger$ \\
SLERP~\cite{ricci2026slerp} & 59.7 & 42.5 & 84.2 & 66.3 & 62.4 & 29.7 & 20.5 & 57.9 & 41.8 & 50.9 \\
LST~\cite{maiorca2023lst} & 56.5 & 39.1 & 79.6 & 62.9 & 57.9 & 25.7 & 17.3 & 51.4 & 37.0 & 44.4 \\
$\lambda$-orth~\cite{ricci2025lorth} & 49.3 & 37.3 & 78.6 & 61.3 & 59.2 & 24.0 & 16.7 & 52.6 & 36.0 & 45.8 \\
\rowcolor{ourblue} \ours{} (ours) & \textbf{61.3} & \textbf{43.3} & \textbf{84.4} & \textbf{66.5} & \textbf{65.5} & \textbf{30.0} & \textbf{21.0} & \textbf{58.9} & \textbf{42.5} & \textbf{51.7} \\
\bottomrule
\end{tabularx}
\end{table*}

\ours is the best compatible method in all 50 combinations of model pair, dataset and direction (Table~\ref{tab:main} and Tables~S1 and~S2 of the supplementary material). It leads the strongest other compatible method by 0.1 to 3.6 points from image to text and by 0.3 to 5.6 points from text to image, and Old $\to$ old by 1.5 to 18.0 points. This includes SLERP, which runs two encoders per query, and XBT, which also adapts the new encoder.

\ours closes much of the gap to re-encoding the gallery. On the two CLIP upgrades it matches or exceeds New $\to$ new in 16 of 20 cells, including 7 of the 10 cells of CLIP ViT-B/32 $\to$ CLIP ViT-L/14 in Table~\ref{tab:main}, although New $\to$ new reencodes the whole gallery. On the three larger upgrades New $\to$ new stays ahead by 0.3 to 21.5 points; \ours recovers 45\% to 95\% of the image to text gain that re-encoding brings over Old $\to$ old, and 9\% to 69\% of the text to image gain, since text queries use no student. Pooled over all settings, \ours obtains 55.0\% of the update gain of BCT~\cite{shen2020bct}, against 36.1\% for SLERP (Table~S9 of the supplementary material).

\subsection{Latency}
\label{sec:latency}
\begin{figure*}[t]
\centering
\includegraphics[width=\textwidth]{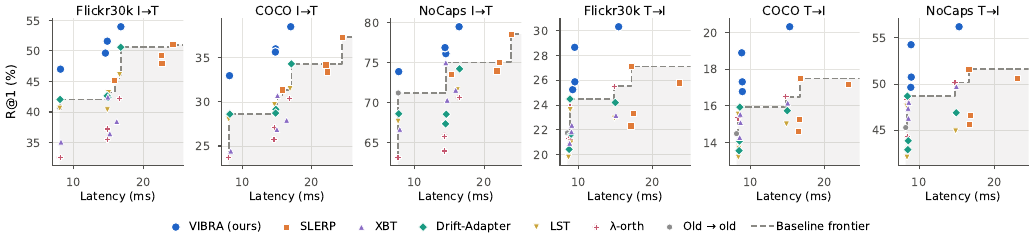}
\caption{Recall@1 against query latency at batch size 1. Each marker is one new backbone, and the dashed line is the frontier of the baselines. Points toward the upper left are better.}
\label{fig:recall_latency}
\end{figure*}

\ours lies above the frontier of the baselines on every dataset and in both directions (Figure~\ref{fig:recall_latency}). At equal or lower latency, \ours is ahead by 3.4 to 9.0 points from image to text and by 0.8 to 6.1 points from text to image.

The latency of \ours is that of the new encoder. It stays within 0.5 ms of the new encoder for image queries, where the low rank update is merged, and is 0.4 to 0.9 ms above it for text queries, which add the text map, while SLERP adds the old encoder and costs 7.4 to 8.2 ms more per query than \ours. The gallery size matters little: NoCaps, whose galleries are a quarter of the size of those of COCO, is faster by about 0.5 ms. Table~S3 of the supplementary material gives latency and throughput for all five pairs. Training \ours takes less than one A100 hour, 19 to 27 times less than XBT and as much as re about two million images (Table~S8 of the supplementary material).

\subsection{Consensus Analysis}
\label{sec:analysis}
\begin{table}[t]
\centering
\footnotesize
\setlength{\tabcolsep}{1.8pt}
\caption{Consensus correction split by the condition of Proposition~\ref{prop:margin} (image $\to$ text, one positive caption per image). Errors corrected: share of the errors of the new view that the barycenter corrects. Columns name the new model for CLIP ViT-B/32; MC$\to$SO: MetaCLIP ViT-B/32 $\to$ SigLIP2 SO400M/14.}
\label{tab:consensus}
\begin{tabular}{lccccc}
\toprule
& L/14 & B/16 & SO400M & MC L/14 & MC$\to$SO \\
\midrule
Queries with $D>0$ (\%) & 91.2 & 90.1 & 93.0 & 92.0 & 88.4 \\
Errors corrected, $D>0$ (\%) & 11.3 & 10.9 & 13.4 & 12.5 & 10.0 \\
Errors corrected, $D=0$ (\%) & 4.2 & 4.2 & 4.5 & 4.2 & 4.0 \\
Net R@1 gain (points) & 4.7 & 4.6 & 5.2 & 5.3 & 1.2 \\
\bottomrule
\end{tabular}
\end{table}

The barycenter corrects errors where Proposition~\ref{prop:margin} locates the dispersion gain (Table~\ref{tab:consensus}). We split the test queries by the condition of the proposition, whether the three views share their hardest irrelevant item. The dispersion gain $D$ is positive on 88.4\% to 93.0\% of the queries, and on these queries the barycenter corrects 10.0\% to 13.4\% of the errors of the new view, 2.5 to 3 times as often as on the queries with $D=0$.

The condition of the proposition is exact on the test sets. On the queries whose views share their hardest irrelevant item the measured $D$ is zero up to numerical precision, and when all three views are wrong on such a query the barycenter corrects none of them, the case of Figure~\ref{fig:sphere}(a). Overall the barycenter raises Recall@1 over the new view on every pair, by 1.2 to 5.3 points. Section~S5 of the supplementary material gives further statistics.

\subsection{Ablation}
\label{sec:ablation}
\begin{table}[t]
\centering
\footnotesize
\setlength{\tabcolsep}{1.8pt}
\caption{Ablation of the components (image $\to$ text R@1, \%); each row adds one component. Columns as in Table~\ref{tab:consensus}.}
\label{tab:components}
\begin{tabular}{lcccccc}
\toprule
& L/14 & B/16 & SO400M & MC L/14 & MC$\to$SO & Mean \\
\midrule
Procrustes map & 48.8 & 48.6 & 46.8 & 50.9 & 56.3 & 50.3 \\
+ Relevance matching & 52.4 & 51.2 & 57.6 & 53.3 & 63.8 & 55.7 \\
+ Consensus teacher & 54.8 & 52.6 & 58.6 & 55.2 & 64.1 & 57.1 \\
\rowcolor{ourblue} \ours (all three) & 55.8 & 53.2 & 58.8 & 55.8 & 64.2 & 57.6 \\
\bottomrule
\end{tabular}
\end{table}

Each component of \ours improves on the previous one, and relevance matching contributes most (Table~\ref{tab:components}, mean over the five datasets). Starting from the Procrustes map, which imitates the old query, relevance matching raises Recall@1 by 2.4 to 10.8 points, 5.4 on average; training the new encoder on the consensus teacher adds 0.3 to 2.4 points, and the auxiliary term another 0.1 to 1.0 points.

Every component helps on every model pair. Both estimates of relevance are needed, and the teacher of three views exceeds every single view and the teacher of two views (Tables~S4 and~S7 of the supplementary material).

\subsection{Sensitivity}
\label{sec:sensitivity}
\begin{figure}[t]
\centering
\includegraphics[width=\linewidth]{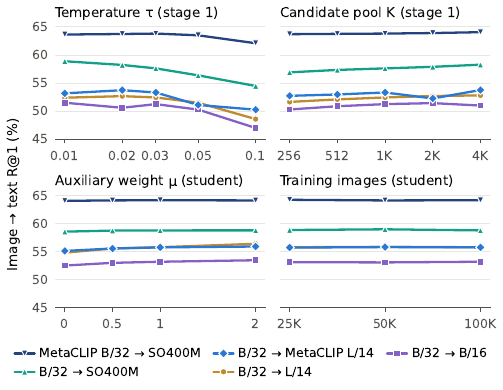}
\caption{Sensitivity of image $\to$ text R@1 to one hyperparameter at a time.}
\label{fig:sensitivity}
\end{figure}

The stage 1 query is insensitive to its hyperparameters around the deployed values. We vary one hyperparameter at a time around the deployed setting (Figure~\ref{fig:sensitivity}, mean over the five datasets). Temperatures from 0.01 to 0.03 stay within 1.3 points and only $\tau=0.1$ loses 1.7 to 4.2, and over $K$ from 256 to 4096 Recall@1 varies by at most 1.5 points. Every $\beta$ between 0.5 and 0.75 is within 1.4 points of the best (Table~S5 of the supplementary material), and the minimum variance weight $\lambda^*$, computed on COCO DEV without tuning, falls at 0.81 to 0.84 and stays within 0.3 points of the best $\lambda$ (Table~S4).

The student is insensitive to the auxiliary weight, the rank of the update and the amount of training data. Doubling $\mu$ changes the student by at most 0.6 points. Ranks 4 and 16 are within 0.1 points of rank 8, so a larger update brings no gain, and students trained on 25K or 50K images with the same number of steps are within 0.2 points of the one trained on 100K.

\section{Conclusion}
\label{sec:conclusion}

\ours upgrades the query encoder of a cross-modal retrieval system without re-encoding the gallery and with one encoder per query. It derives its training target from the retrieval objective, pools the views of several models into a barycenter that a single query reproduces, and transfers this target into the new encoder. On five benchmarks with five upgrades, it is the best compatible method in all 50 settings, improving Recall@1 by up to 5.6 points over the strongest baseline and by up to 18.0 points over the old model, with less than 1 ms added to the latency of the new encoder.

\section*{LLM Usage Statement}
In this work, we used a large language model assistant to implement the method, the baselines and the evaluation in code, to design and give feedback on experiments, including the preregistered sensitivity and ablation studies, to check derivations and assist in writing the proofs of the supplementary material, and to help interpret the results. We have not used it to generate synthetic data sets, and translation and qualitative data analysis are not applicable to this work. Additionally, we used it to suggest experimental parameters, to create and modify the figures, to search for and summarize related work, to draft and edit parts of the text.

We have reviewed all AI assisted work. The authors defined the problem and the method, inspected the code and ran every experiment, checked each reported number against the experiment outputs, verified the derivations and proofs by hand, and read the cited papers to confirm the statements made about them. We take responsibility for the final content of this work, including text, claims and artifacts produced with the aid of generative AI.

{
    \small
    \bibliographystyle{ieeenat_fullname}
    \bibliography{main}
}

\clearpage
\appendix
\setcounter{section}{0}\setcounter{table}{0}\setcounter{figure}{0}
\renewcommand{\thesection}{S\arabic{section}}\renewcommand{\theHsection}{S\arabic{section}}
\renewcommand{\thetable}{S\arabic{table}}\renewcommand{\theHtable}{S\arabic{table}}
\renewcommand{\thefigure}{S\arabic{figure}}\renewcommand{\theHfigure}{S\arabic{figure}}
\twocolumn[\begin{center}{\Large\bf Supplementary Material}\end{center}\vspace{1em}]

Section, equation and proposition numbers without the prefix S refer to the main paper.

\section{Full Retrieval Results}
\label{sec:supp_full}
Tables~\ref{tab:supp_i2t} and~\ref{tab:supp_t2i} extend the R@1 results of the main paper to R@5 and R@10 for every model pair, dataset and method.
\begin{table*}[t]
\centering
\scriptsize
\setlength{\tabcolsep}{2.6pt}
\caption{Image $\to$ text retrieval against the old gallery (R@1 / R@5 / R@10, \%). The old gallery is never reencoded. Every method except SLERP encodes the query with one network; SLERP runs the old and the new encoder. Gray rows are references: Old $\to$ old keeps the old model, New $\to$ new reencodes the whole gallery. $^\dagger$ published number. $^\ddagger$ our reproduction of XBT at the scale of its paper, used where the paper reports no number. \textbf{Bold} marks the best compatible method.}
\label{tab:supp_i2t}
\resizebox{\textwidth}{!}{%
\begin{tabular}{lrrrrrrrrrrrrrrr}
\toprule
Method & \multicolumn{3}{c}{Flickr30k} & \multicolumn{3}{c}{COCO} & \multicolumn{3}{c}{NoCaps} & \multicolumn{3}{c}{CxC} & \multicolumn{3}{c}{XM3600} \\
\cmidrule(lr){2-4} \cmidrule(lr){5-7} \cmidrule(lr){8-10} \cmidrule(lr){11-13} \cmidrule(lr){14-16}
& R@1 & R@5 & R@10 & R@1 & R@5 & R@10 & R@1 & R@5 & R@10 & R@1 & R@5 & R@10 & R@1 & R@5 & R@10 \\
\midrule
\multicolumn{16}{l}{\textit{CLIP ViT-B/32 $\to$ CLIP ViT-L/14}} \\
\rowcolor{refgray} Old $\to$ old & 40.7 & 64.7 & 73.7 & 28.6 & 50.4 & 59.9 & 71.2 & 91.9 & 96.3 & 51.6 & 77.4 & 85.5 & 53.3 & 79.1 & 86.5 \\
\rowcolor{refgray} New $\to$ new & 48.7 & 73.0 & 81.2 & 34.4 & 56.0 & 65.4 & 73.4 & 93.5 & 97.4 & 58.0 & 81.7 & 88.4 & 54.1 & 78.8 & 86.9 \\
Drift-Adapter~\cite{vejendla2025drift} & 42.4 & 66.8 & 76.0 & 29.1 & 50.1 & 59.3 & 67.4 & 90.1 & 95.4 & 52.4 & 76.8 & 84.8 & 45.9 & 72.6 & 81.8 \\
XBT~\cite{jang2025xbt} & 42.5$^\dagger$ & 66.0$^\dagger$ & 74.9$^\dagger$ & 30.7$^\dagger$ & 52.6$^\dagger$ & 62.3$^\dagger$ & 75.0$^\dagger$ & 93.3$^\dagger$ & 97.3$^\dagger$ & 50.8$^\ddagger$ & 76.0$^\ddagger$ & 85.0$^\ddagger$ & 50.8$^\ddagger$ & 77.2$^\ddagger$ & 85.6$^\ddagger$ \\
SLERP~\cite{ricci2026slerp} & 48.0$^\dagger$ & 71.8$^\dagger$ & 80.3$^\dagger$ & 33.4$^\dagger$ & 55.4$^\dagger$ & 64.8$^\dagger$ & 74.0$^\dagger$ & 93.3$^\dagger$ & 96.9$^\dagger$ & 56.0 & 80.6 & 88.1 & 55.6 & 81.3 & 88.2 \\
LST~\cite{maiorca2023lst} & 40.4 & 64.9 & 74.2 & 28.6 & 49.5 & 58.9 & 67.3 & 89.4 & 95.3 & 52.4 & 76.5 & 85.0 & 45.9 & 72.4 & 81.9 \\
$\lambda$-orth~\cite{ricci2025lorth} & 35.6 & 59.6 & 69.6 & 25.7 & 46.4 & 56.0 & 64.0 & 87.9 & 94.1 & 49.5 & 74.5 & 84.1 & 47.2 & 74.3 & 83.5 \\
\rowcolor{ourblue} \ours{} (ours) & \textbf{51.6} & \textbf{75.7} & \textbf{83.7} & \textbf{36.0} & \textbf{58.6} & \textbf{67.9} & \textbf{76.1} & \textbf{94.4} & \textbf{97.9} & \textbf{59.6} & \textbf{83.0} & \textbf{90.1} & \textbf{56.1} & \textbf{81.9} & \textbf{89.4} \\
\midrule
\multicolumn{16}{l}{\textit{CLIP ViT-B/32 $\to$ CLIP ViT-B/16}} \\
\rowcolor{refgray} Old $\to$ old & 40.7 & 64.7 & 73.7 & 28.6 & 50.4 & 59.9 & 71.2 & 91.9 & 96.3 & 51.6 & 77.4 & 85.5 & 53.3 & 79.1 & 86.5 \\
\rowcolor{refgray} New $\to$ new & 44.4 & 68.6 & 77.4 & 30.9 & 52.4 & 62.0 & 71.8 & 91.9 & 96.8 & 54.1 & 78.4 & 86.9 & 55.1 & 80.1 & 87.6 \\
Drift-Adapter~\cite{vejendla2025drift} & 42.1 & 66.4 & 75.7 & 28.6 & 49.8 & 59.2 & 68.6 & 90.7 & 95.5 & 52.1 & 77.0 & 85.4 & 49.6 & 76.6 & 85.3 \\
XBT~\cite{jang2025xbt} & 35.1$^\ddagger$ & 58.2$^\ddagger$ & 67.9$^\ddagger$ & 24.4$^\ddagger$ & 44.6$^\ddagger$ & 54.2$^\ddagger$ & 66.7$^\ddagger$ & 89.9$^\ddagger$ & 95.0$^\ddagger$ & 47.2$^\ddagger$ & 73.3$^\ddagger$ & 82.6$^\ddagger$ & 48.3$^\ddagger$ & 75.9$^\ddagger$ & 84.6$^\ddagger$ \\
SLERP~\cite{ricci2026slerp} & 45.2 & 69.8 & 78.5 & 31.4 & 53.4 & 62.6 & 73.6 & \textbf{93.1} & \textbf{97.0} & 55.4 & 79.5 & 87.2 & 54.7 & 81.0 & 87.8 \\
LST~\cite{maiorca2023lst} & 40.6 & 64.6 & 74.2 & 28.0 & 48.9 & 58.5 & 67.7 & 90.3 & 95.6 & 51.1 & 76.8 & 85.0 & 49.8 & 75.6 & 84.5 \\
$\lambda$-orth~\cite{ricci2025lorth} & 32.6 & 56.2 & 66.4 & 23.7 & 44.0 & 53.5 & 63.2 & 87.2 & 93.4 & 46.5 & 72.9 & 82.2 & 47.5 & 74.7 & 83.5 \\
\rowcolor{ourblue} \ours{} (ours) & \textbf{47.0} & \textbf{71.9} & \textbf{80.2} & \textbf{33.0} & \textbf{55.1} & \textbf{64.5} & \textbf{73.9} & 92.7 & 96.9 & \textbf{57.0} & \textbf{81.1} & \textbf{88.1} & \textbf{54.8} & \textbf{81.7} & \textbf{89.5} \\
\midrule
\multicolumn{16}{l}{\textit{CLIP ViT-B/32 $\to$ SigLIP2 SO400M/14}} \\
\rowcolor{refgray} Old $\to$ old & 40.7 & 64.7 & 73.7 & 28.6 & 50.4 & 59.9 & 71.2 & 91.9 & 96.3 & 51.6 & 77.4 & 85.5 & 53.3 & 79.1 & 86.5 \\
\rowcolor{refgray} New $\to$ new & 69.1 & 88.4 & 93.1 & 50.4 & 72.6 & 80.1 & 88.5 & 98.5 & 99.6 & 72.4 & 90.7 & 95.3 & 65.8 & 87.5 & 93.0 \\
Drift-Adapter~\cite{vejendla2025drift} & 50.6 & 74.1 & 81.9 & 34.3 & 56.4 & 65.5 & 74.2 & 93.4 & 97.3 & 58.2 & 81.8 & 88.6 & 52.9 & 78.0 & 86.0 \\
XBT~\cite{jang2025xbt} & 38.5$^\ddagger$ & 62.0$^\ddagger$ & 71.4$^\ddagger$ & 27.9$^\ddagger$ & 48.9$^\ddagger$ & 59.0$^\ddagger$ & 71.6$^\ddagger$ & 92.8$^\ddagger$ & 96.7$^\ddagger$ & 51.8$^\ddagger$ & 77.6$^\ddagger$ & 86.0$^\ddagger$ & 54.2$^\ddagger$ & 80.1$^\ddagger$ & 87.2$^\ddagger$ \\
SLERP~\cite{ricci2026slerp} & 51.0$^\dagger$ & 75.3$^\dagger$ & 83.5$^\dagger$ & 37.4$^\dagger$ & 60.4$^\dagger$ & 69.3$^\dagger$ & 78.6$^\dagger$ & 95.3$^\dagger$ & \textbf{98.3$^\dagger$} & 61.7 & 84.2 & 91.0 & 58.9 & 83.9 & 90.5 \\
LST~\cite{maiorca2023lst} & 46.1 & 70.0 & 78.5 & 31.5 & 53.4 & 62.8 & 71.6 & 91.7 & 96.0 & 55.3 & 79.6 & 87.2 & 49.9 & 76.1 & 84.3 \\
$\lambda$-orth~\cite{ricci2025lorth} & 42.2 & 66.7 & 76.1 & 30.4 & 52.0 & 61.6 & 70.6 & 91.9 & 96.2 & 55.4 & 80.1 & 87.7 & 53.8 & 79.5 & 87.2 \\
\rowcolor{ourblue} \ours{} (ours) & \textbf{54.0} & \textbf{77.7} & \textbf{85.2} & \textbf{38.5} & \textbf{61.4} & \textbf{70.4} & \textbf{79.5} & \textbf{95.5} & 98.2 & \textbf{62.2} & \textbf{84.5} & \textbf{91.4} & \textbf{60.8} & \textbf{84.8} & \textbf{91.4} \\
\midrule
\multicolumn{16}{l}{\textit{CLIP ViT-B/32 $\to$ MetaCLIP ViT-L/14}} \\
\rowcolor{refgray} Old $\to$ old & 40.7 & 64.7 & 73.7 & 28.6 & 50.4 & 59.9 & 71.2 & 91.9 & 96.3 & 51.6 & 77.4 & 85.5 & 53.3 & 79.1 & 86.5 \\
\rowcolor{refgray} New $\to$ new & 57.1 & 80.3 & 87.6 & 42.4 & 64.9 & 73.5 & 82.4 & 96.8 & 99.0 & 65.5 & 86.5 & 92.3 & 62.2 & 85.8 & 91.8 \\
Drift-Adapter~\cite{vejendla2025drift} & 42.6 & 66.6 & 75.9 & 28.7 & 49.8 & 59.3 & 68.6 & 90.6 & 95.6 & 51.6 & 76.6 & 84.8 & 47.0 & 72.7 & 81.4 \\
XBT~\cite{jang2025xbt} & 36.5$^\ddagger$ & 60.2$^\ddagger$ & 69.9$^\ddagger$ & 26.9$^\ddagger$ & 47.4$^\ddagger$ & 57.1$^\ddagger$ & 70.3$^\ddagger$ & 91.8$^\ddagger$ & 96.4$^\ddagger$ & 50.6$^\ddagger$ & 77.0$^\ddagger$ & 85.0$^\ddagger$ & 53.1$^\ddagger$ & 79.1$^\ddagger$ & 86.6$^\ddagger$ \\
SLERP~\cite{ricci2026slerp} & 49.2 & 73.2 & 81.4 & 34.2 & 56.5 & 65.7 & 75.1 & \textbf{93.7} & 97.2 & 57.9 & 81.2 & 88.7 & 55.3 & 80.3 & 88.1 \\
LST~\cite{maiorca2023lst} & 43.2 & 67.4 & 76.3 & 29.7 & 51.2 & 60.6 & 68.4 & 90.2 & 95.4 & 52.8 & 77.7 & 86.1 & 47.4 & 74.1 & 82.7 \\
$\lambda$-orth~\cite{ricci2025lorth} & 37.2 & 61.8 & 71.7 & 27.1 & 48.3 & 57.8 & 65.8 & 89.4 & 94.9 & 51.5 & 76.3 & 85.3 & 48.5 & 76.5 & 84.8 \\
\rowcolor{ourblue} \ours{} (ours) & \textbf{49.6} & \textbf{73.8} & \textbf{81.9} & \textbf{35.7} & \textbf{58.2} & \textbf{67.7} & \textbf{76.9} & 93.7 & \textbf{97.6} & \textbf{59.3} & \textbf{82.8} & \textbf{89.8} & \textbf{57.6} & \textbf{82.9} & \textbf{89.6} \\
\midrule
\multicolumn{16}{l}{\textit{MetaCLIP ViT-B/32 $\to$ SigLIP2 SO400M/14}} \\
\rowcolor{refgray} Old $\to$ old & 43.3 & 67.2 & 76.0 & 33.0 & 55.3 & 64.7 & 77.9 & 94.5 & 97.6 & 56.7 & 80.8 & 88.4 & 59.2 & 83.9 & 90.7 \\
\rowcolor{refgray} New $\to$ new & 69.1 & 88.4 & 93.1 & 50.4 & 72.6 & 80.1 & 88.5 & 98.5 & 99.6 & 72.4 & 90.7 & 95.3 & 65.8 & 87.5 & 93.0 \\
Drift-Adapter~\cite{vejendla2025drift} & 58.5 & 80.9 & 87.6 & 40.9 & 63.5 & 72.2 & 81.0 & 95.3 & 98.0 & 64.5 & 86.1 & 92.0 & 58.8 & 83.6 & 90.1 \\
XBT~\cite{jang2025xbt} & 49.0$^\ddagger$ & 72.1$^\ddagger$ & 80.4$^\ddagger$ & 34.4$^\ddagger$ & 56.7$^\ddagger$ & 65.8$^\ddagger$ & 79.8$^\ddagger$ & 95.5$^\ddagger$ & 98.4$^\ddagger$ & 59.0$^\ddagger$ & 82.6$^\ddagger$ & 89.4$^\ddagger$ & 57.8$^\ddagger$ & 81.8$^\ddagger$ & 87.9$^\ddagger$ \\
SLERP~\cite{ricci2026slerp} & 59.7 & 81.2 & 87.8 & 42.5 & 65.0 & 73.7 & 84.2 & 97.3 & 99.0 & 66.3 & 87.4 & \textbf{93.2} & 62.4 & 86.5 & 92.1 \\
LST~\cite{maiorca2023lst} & 56.5 & 79.2 & 86.5 & 39.1 & 61.8 & 70.7 & 79.6 & 95.1 & 97.8 & 62.9 & 84.5 & 91.2 & 57.9 & 82.9 & 90.0 \\
$\lambda$-orth~\cite{ricci2025lorth} & 49.3 & 73.2 & 81.8 & 37.3 & 59.7 & 69.0 & 78.6 & 94.7 & 97.7 & 61.3 & 84.4 & 90.9 & 59.2 & 84.3 & 91.1 \\
\rowcolor{ourblue} \ours{} (ours) & \textbf{61.3} & \textbf{83.1} & \textbf{89.3} & \textbf{43.3} & \textbf{65.9} & \textbf{74.2} & \textbf{84.4} & \textbf{97.5} & \textbf{99.2} & \textbf{66.5} & \textbf{87.8} & 92.9 & \textbf{65.5} & \textbf{88.6} & \textbf{93.7} \\
\bottomrule
\end{tabular}}
\end{table*}
\begin{table*}[t]
\centering
\scriptsize
\setlength{\tabcolsep}{2.6pt}
\caption{Text $\to$ image retrieval against the old gallery (R@1 / R@5 / R@10, \%). The old gallery is never reencoded. Every method except SLERP encodes the query with one network; SLERP runs the old and the new encoder. Gray rows are references: Old $\to$ old keeps the old model, New $\to$ new reencodes the whole gallery. $^\dagger$ published number. $^\ddagger$ our reproduction of XBT at the scale of its paper, used where the paper reports no number. \textbf{Bold} marks the best compatible method.}
\label{tab:supp_t2i}
\resizebox{\textwidth}{!}{%
\begin{tabular}{lrrrrrrrrrrrrrrr}
\toprule
Method & \multicolumn{3}{c}{Flickr30k} & \multicolumn{3}{c}{COCO} & \multicolumn{3}{c}{NoCaps} & \multicolumn{3}{c}{CxC} & \multicolumn{3}{c}{XM3600} \\
\cmidrule(lr){2-4} \cmidrule(lr){5-7} \cmidrule(lr){8-10} \cmidrule(lr){11-13} \cmidrule(lr){14-16}
& R@1 & R@5 & R@10 & R@1 & R@5 & R@10 & R@1 & R@5 & R@10 & R@1 & R@5 & R@10 & R@1 & R@5 & R@10 \\
\midrule
\multicolumn{16}{l}{\textit{CLIP ViT-B/32 $\to$ CLIP ViT-L/14}} \\
\rowcolor{refgray} Old $\to$ old & 21.8 & 41.6 & 51.1 & 14.5 & 30.2 & 38.9 & 45.3 & 75.0 & 84.6 & 32.3 & 58.8 & 70.0 & 38.7 & 66.5 & 76.5 \\
\rowcolor{refgray} New $\to$ new & 28.2 & 49.6 & 59.0 & 18.7 & 35.9 & 44.4 & 47.8 & 76.6 & 85.2 & 38.2 & 63.9 & 74.2 & 42.5 & 69.2 & 77.9 \\
Drift-Adapter~\cite{vejendla2025drift} & 21.7 & 41.6 & 51.1 & 14.1 & 29.9 & 38.5 & 43.9 & 74.3 & 83.8 & 31.9 & 58.3 & 69.5 & 38.6 & 66.6 & 76.4 \\
XBT~\cite{jang2025xbt} & 22.4$^\dagger$ & 42.7$^\dagger$ & 52.4$^\dagger$ & 15.6$^\dagger$ & 32.3$^\dagger$ & 41.3$^\dagger$ & 48.0$^\dagger$ & 79.0$^\dagger$ & 88.2$^\dagger$ & 33.9$^\ddagger$ & 60.6$^\ddagger$ & 71.8$^\ddagger$ & 41.6$^\ddagger$ & 70.7$^\ddagger$ & 80.0$^\ddagger$ \\
SLERP~\cite{ricci2026slerp} & 23.3$^\dagger$ & 43.8$^\dagger$ & 53.3$^\dagger$ & 15.3$^\dagger$ & 31.6$^\dagger$ & 40.3$^\dagger$ & 46.6$^\dagger$ & 76.5$^\dagger$ & 85.5$^\dagger$ & 33.8 & 60.4 & 71.5 & 40.5 & 68.9 & 78.6 \\
LST~\cite{maiorca2023lst} & 20.9 & 40.4 & 49.9 & 13.7 & 29.1 & 37.7 & 42.9 & 73.3 & 83.2 & 30.9 & 57.5 & 68.8 & 37.1 & 65.5 & 75.8 \\
$\lambda$-orth~\cite{ricci2025lorth} & 21.2 & 40.6 & 50.0 & 13.8 & 29.1 & 37.5 & 44.4 & 74.5 & 84.2 & 31.4 & 57.4 & 68.5 & 40.3 & 68.0 & 77.6 \\
\rowcolor{ourblue} \ours{} (ours) & \textbf{25.9} & \textbf{47.2} & \textbf{56.9} & \textbf{17.3} & \textbf{34.9} & \textbf{44.0} & \textbf{50.8} & \textbf{80.2} & \textbf{88.6} & \textbf{36.8} & \textbf{63.8} & \textbf{74.4} & \textbf{45.1} & \textbf{73.3} & \textbf{81.3} \\
\midrule
\multicolumn{16}{l}{\textit{CLIP ViT-B/32 $\to$ CLIP ViT-B/16}} \\
\rowcolor{refgray} Old $\to$ old & 21.8 & 41.6 & 51.1 & 14.5 & 30.2 & 38.9 & 45.3 & 75.0 & 84.6 & 32.3 & 58.8 & 70.0 & 38.7 & 66.5 & 76.5 \\
\rowcolor{refgray} New $\to$ new & 24.9 & 45.4 & 54.9 & 16.5 & 33.0 & 41.6 & 46.8 & 76.1 & 85.0 & 34.6 & 61.1 & 72.0 & 42.2 & 68.5 & 77.0 \\
Drift-Adapter~\cite{vejendla2025drift} & 20.4 & 39.8 & 49.2 & 13.6 & 28.7 & 37.1 & 42.9 & 72.9 & 83.0 & 30.6 & 57.3 & 68.8 & 36.3 & 64.3 & 74.4 \\
XBT~\cite{jang2025xbt} & 20.9$^\ddagger$ & 40.7$^\ddagger$ & 50.6$^\ddagger$ & 14.3$^\ddagger$ & 30.5$^\ddagger$ & 39.4$^\ddagger$ & 46.3$^\ddagger$ & 77.7$^\ddagger$ & 87.2$^\ddagger$ & 32.9$^\ddagger$ & 59.7$^\ddagger$ & 71.0$^\ddagger$ & 41.2$^\ddagger$ & 70.1$^\ddagger$ & 79.8$^\ddagger$ \\
SLERP~\cite{ricci2026slerp} & 22.3 & 42.4 & 51.8 & 14.6 & 30.4 & 39.2 & 45.6 & 75.3 & 84.7 & 32.4 & 59.3 & 70.3 & 38.6 & 66.2 & 76.7 \\
LST~\cite{maiorca2023lst} & 19.8 & 38.9 & 48.3 & 13.2 & 28.2 & 36.4 & 42.1 & 72.1 & 82.3 & 30.0 & 56.6 & 68.3 & 35.5 & 63.5 & 73.7 \\
$\lambda$-orth~\cite{ricci2025lorth} & 21.3 & 40.7 & 50.1 & 13.8 & 29.1 & 37.5 & 44.2 & 74.4 & 84.1 & 31.5 & 57.3 & 68.5 & 40.0 & 67.9 & 77.6 \\
\rowcolor{ourblue} \ours{} (ours) & \textbf{25.2} & \textbf{46.5} & \textbf{56.1} & \textbf{16.8} & \textbf{34.2} & \textbf{43.3} & \textbf{49.6} & \textbf{79.5} & \textbf{88.0} & \textbf{36.4} & \textbf{63.3} & \textbf{74.0} & \textbf{43.9} & \textbf{72.2} & \textbf{80.8} \\
\midrule
\multicolumn{16}{l}{\textit{CLIP ViT-B/32 $\to$ SigLIP2 SO400M/14}} \\
\rowcolor{refgray} Old $\to$ old & 21.8 & 41.6 & 51.1 & 14.5 & 30.2 & 38.9 & 45.3 & 75.0 & 84.6 & 32.3 & 58.8 & 70.0 & 38.7 & 66.5 & 76.5 \\
\rowcolor{refgray} New $\to$ new & 51.5 & 73.2 & 80.4 & 35.1 & 56.5 & 65.0 & 70.1 & 91.0 & 95.1 & 56.8 & 80.2 & 87.3 & 58.8 & 82.6 & 89.2 \\
Drift-Adapter~\cite{vejendla2025drift} & 24.2 & 44.6 & 54.3 & 15.7 & 32.6 & 41.4 & 46.9 & 76.0 & 84.9 & 34.7 & 61.2 & 72.1 & 39.9 & 68.3 & 77.8 \\
XBT~\cite{jang2025xbt} & 23.2$^\ddagger$ & 44.2$^\ddagger$ & 54.2$^\ddagger$ & 16.2$^\ddagger$ & 33.4$^\ddagger$ & 42.6$^\ddagger$ & 49.7$^\ddagger$ & 80.9$^\ddagger$ & 89.4$^\ddagger$ & 35.7$^\ddagger$ & 62.9$^\ddagger$ & 73.7$^\ddagger$ & 43.4$^\ddagger$ & 72.6$^\ddagger$ & 81.6$^\ddagger$ \\
SLERP~\cite{ricci2026slerp} & 25.8$^\dagger$ & 47.2$^\dagger$ & 56.8$^\dagger$ & 17.2$^\dagger$ & 34.8$^\dagger$ & 44.0$^\dagger$ & 50.6$^\dagger$ & 79.7$^\dagger$ & 87.6$^\dagger$ & 36.5 & 63.5 & 74.4 & 43.2 & 71.4 & 80.4 \\
LST~\cite{maiorca2023lst} & 22.9 & 43.1 & 52.6 & 15.0 & 31.2 & 39.9 & 44.9 & 73.9 & 83.2 & 33.4 & 59.6 & 70.5 & 38.4 & 66.6 & 76.8 \\
$\lambda$-orth~\cite{ricci2025lorth} & 25.5 & 46.1 & 55.6 & 16.5 & 33.5 & 42.4 & 50.2 & 78.5 & 87.1 & 35.7 & 62.1 & 72.5 & 43.6 & 71.3 & 80.3 \\
\rowcolor{ourblue} \ours{} (ours) & \textbf{30.3} & \textbf{52.8} & \textbf{62.4} & \textbf{20.3} & \textbf{39.3} & \textbf{48.6} & \textbf{56.2} & \textbf{84.0} & \textbf{91.1} & \textbf{41.2} & \textbf{68.0} & \textbf{77.9} & \textbf{48.7} & \textbf{76.6} & \textbf{84.8} \\
\midrule
\multicolumn{16}{l}{\textit{CLIP ViT-B/32 $\to$ MetaCLIP ViT-L/14}} \\
\rowcolor{refgray} Old $\to$ old & 21.8 & 41.6 & 51.1 & 14.5 & 30.2 & 38.9 & 45.3 & 75.0 & 84.6 & 32.3 & 58.8 & 70.0 & 38.7 & 66.5 & 76.5 \\
\rowcolor{refgray} New $\to$ new & 40.9 & 63.8 & 72.3 & 26.9 & 47.3 & 56.4 & 60.9 & 86.2 & 92.3 & 48.7 & 73.6 & 82.5 & 52.4 & 78.2 & 85.7 \\
Drift-Adapter~\cite{vejendla2025drift} & 24.5 & 45.3 & 55.0 & 15.9 & 33.0 & 42.1 & 48.7 & 78.3 & 86.9 & 35.0 & 62.0 & 73.2 & 42.4 & 71.4 & 80.6 \\
XBT~\cite{jang2025xbt} & 21.9$^\ddagger$ & 42.2$^\ddagger$ & 52.2$^\ddagger$ & 15.1$^\ddagger$ & 31.7$^\ddagger$ & 40.7$^\ddagger$ & 47.4$^\ddagger$ & 78.8$^\ddagger$ & 88.1$^\ddagger$ & 33.9$^\ddagger$ & 61.0$^\ddagger$ & 72.4$^\ddagger$ & 42.8$^\ddagger$ & 71.8$^\ddagger$ & 81.3$^\ddagger$ \\
SLERP~\cite{ricci2026slerp} & 27.1 & 48.7 & 58.4 & 17.5 & 35.2 & 44.4 & 51.6 & 80.7 & 88.8 & 37.3 & 64.3 & 75.0 & 44.8 & 73.2 & 81.8 \\
LST~\cite{maiorca2023lst} & 23.6 & 44.1 & 53.6 & 15.5 & 32.3 & 41.2 & 47.6 & 77.3 & 86.2 & 34.4 & 61.3 & 72.5 & 42.2 & 70.5 & 79.9 \\
$\lambda$-orth~\cite{ricci2025lorth} & 23.8 & 44.4 & 53.9 & 15.4 & 31.9 & 40.7 & 48.3 & 77.8 & 86.7 & 34.5 & 60.9 & 71.4 & 42.4 & 71.5 & 80.4 \\
\rowcolor{ourblue} \ours{} (ours) & \textbf{28.6} & \textbf{50.8} & \textbf{60.5} & \textbf{18.9} & \textbf{37.6} & \textbf{47.0} & \textbf{54.3} & \textbf{82.8} & \textbf{90.2} & \textbf{39.6} & \textbf{66.6} & \textbf{77.2} & \textbf{48.1} & \textbf{76.5} & \textbf{84.5} \\
\midrule
\multicolumn{16}{l}{\textit{MetaCLIP ViT-B/32 $\to$ SigLIP2 SO400M/14}} \\
\rowcolor{refgray} Old $\to$ old & 28.0 & 49.7 & 59.1 & 19.3 & 37.6 & 47.0 & 55.9 & 83.6 & 90.7 & 40.1 & 67.0 & 77.1 & 49.9 & 76.2 & 84.3 \\
\rowcolor{refgray} New $\to$ new & 51.5 & 73.2 & 80.4 & 35.1 & 56.5 & 65.0 & 70.1 & 91.0 & 95.1 & 56.8 & 80.2 & 87.3 & 58.8 & 82.6 & 89.2 \\
Drift-Adapter~\cite{vejendla2025drift} & 26.8 & 48.0 & 57.5 & 18.2 & 36.3 & 45.6 & 53.3 & 81.0 & 88.8 & 38.6 & 65.2 & 75.8 & 45.9 & 73.2 & 82.5 \\
XBT~\cite{jang2025xbt} & 22.7$^\ddagger$ & 43.3$^\ddagger$ & 53.0$^\ddagger$ & 17.0$^\ddagger$ & 34.6$^\ddagger$ & 43.8$^\ddagger$ & 53.1$^\ddagger$ & 82.8$^\ddagger$ & 90.7$^\ddagger$ & 37.6$^\ddagger$ & 64.4$^\ddagger$ & 75.4$^\ddagger$ & 45.9$^\ddagger$ & 74.5$^\ddagger$ & 82.8$^\ddagger$ \\
SLERP~\cite{ricci2026slerp} & 29.7 & 51.7 & 61.0 & 20.5 & 39.5 & 48.8 & 57.9 & 84.8 & 91.5 & 41.8 & 68.4 & 78.5 & 50.9 & 77.7 & 85.3 \\
LST~\cite{maiorca2023lst} & 25.7 & 46.7 & 56.0 & 17.3 & 34.8 & 43.7 & 51.4 & 79.4 & 87.4 & 37.0 & 63.5 & 73.9 & 44.4 & 71.9 & 81.1 \\
$\lambda$-orth~\cite{ricci2025lorth} & 24.0 & 44.2 & 53.7 & 16.7 & 33.7 & 42.7 & 52.6 & 80.4 & 88.5 & 36.0 & 62.1 & 73.2 & 45.8 & 73.2 & 82.1 \\
\rowcolor{ourblue} \ours{} (ours) & \textbf{30.0} & \textbf{52.1} & \textbf{61.5} & \textbf{21.0} & \textbf{40.2} & \textbf{49.6} & \textbf{58.9} & \textbf{85.4} & \textbf{92.0} & \textbf{42.5} & \textbf{68.8} & \textbf{79.0} & \textbf{51.7} & \textbf{78.4} & \textbf{85.9} \\
\bottomrule
\end{tabular}}
\end{table*}

\section{Query Latency}
\label{sec:supp_latency}
Table~\ref{tab:supp_latency} lists the latency and throughput of every method on the Large COCO galleries for all five model pairs.
\begin{table*}[t]
\centering
\scriptsize
\setlength{\tabcolsep}{3pt}
\caption{Query latency on the Large COCO galleries: ms per query at batch size 1 / queries per second at batch size 256. One A100-SXM4-40GB, bf16, exact top 10 search over the old gallery (175,680 captions for image $\to$ text, 35,136 images for text $\to$ image); preprocessing and tokenization excluded.}
\label{tab:supp_latency}
\resizebox{\textwidth}{!}{%
\begin{tabular}{llrrrrrrrr}
\toprule
& Pair & Old $\to$ old & New $\to$ new & Drift-Adapter & XBT & SLERP & LST & $\lambda$-orth & \ours{} \\
\midrule
\multirow{5}{*}{I$\to$T} & CLIP ViT-B/32 $\to$ CLIP ViT-L/14 & 9.0 / 8603 & 16.6 / 711 & 16.6 / 710 & 16.8 / 688 & 25.2 / 659 & 16.5 / 709 & 16.5 / 709 & 16.6 / 709 \\
 & CLIP ViT-B/32 $\to$ CLIP ViT-B/16 & 9.0 / 8600 & 8.8 / 2707 & 9.0 / 2702 & 9.1 / 2560 & 17.6 / 2083 & 8.9 / 2693 & 8.9 / 2693 & 8.9 / 2693 \\
 & CLIP ViT-B/32 $\to$ SigLIP2 SO400M/14 & 8.9 / 8588 & 19.0 / 549 & 18.6 / 547 & 18.1 / 590 & 27.1 / 517 & 18.8 / 546 & 18.7 / 546 & 18.9 / 546 \\
 & CLIP ViT-B/32 $\to$ MetaCLIP ViT-L/14 & 8.8 / 8601 & 16.2 / 712 & 16.2 / 711 & 16.4 / 687 & 24.4 / 660 & 16.2 / 710 & 16.3 / 710 & 16.2 / 710 \\
 & MetaCLIP ViT-B/32 $\to$ SigLIP2 SO400M/14 & 8.9 / 8587 & 19.1 / 549 & 19.0 / 547 & 18.3 / 590 & 27.4 / 517 & 19.0 / 546 & 18.9 / 546 & 18.9 / 546 \\
\midrule
\multirow{5}{*}{T$\to$I} & CLIP ViT-B/32 $\to$ CLIP ViT-L/14 & 9.1 / 26898 & 9.2 / 19041 & 9.5 / 17226 & 9.6 / 18954 & 18.4 / 11298 & 9.5 / 19018 & 9.4 / 19099 & 10.0 / 18844 \\
 & CLIP ViT-B/32 $\to$ CLIP ViT-B/16 & 8.9 / 27155 & 8.9 / 27146 & 9.3 / 26123 & 9.3 / 26092 & 18.2 / 13550 & 9.1 / 26869 & 9.1 / 26782 & 9.7 / 25377 \\
 & CLIP ViT-B/32 $\to$ SigLIP2 SO400M/14 & 9.0 / 26657 & 16.7 / 2517 & 16.7 / 2523 & 16.7 / 2522 & 26.1 / 2321 & 16.4 / 2523 & 16.5 / 2524 & 17.0 / 2519 \\
 & CLIP ViT-B/32 $\to$ MetaCLIP ViT-L/14 & 9.1 / 26928 & 9.2 / 19028 & 9.5 / 19066 & 9.4 / 17748 & 18.4 / 11384 & 9.3 / 18866 & 9.3 / 19096 & 9.8 / 18842 \\
 & MetaCLIP ViT-B/32 $\to$ SigLIP2 SO400M/14 & 9.0 / 26886 & 16.5 / 2517 & 16.7 / 2523 & 16.6 / 2521 & 25.8 / 2318 & 16.3 / 2523 & 16.3 / 2524 & 16.9 / 2519 \\
\bottomrule
\end{tabular}}
\end{table*}

\section{Proofs}
\label{sec:supp_proofs}

Throughout, gallery vectors and views are unit vectors, the weights $c$ lie in the simplex, $\bar q=\sum_v c_v q_v$, and $\mathcal{N}$ is a finite set with at least $k$ elements. For a function $\xi$ on $\mathcal{N}$, $\xi_{(k)}$ denotes its $k$th smallest value. We use two elementary facts about order statistics: if $\xi(n)\le\zeta(n)$ for all $n$, then $\xi_{(k)}\le\zeta_{(k)}$; and for every subset $S\subseteq\mathcal{N}$ with $|S|=k$, $\xi_{(k)}\le\max_{n\in S}\xi(n)$.

\subsection{Relevance matching}

\paragraph{Optimality condition.}
With $\ell_y=\langle q,g_y\rangle/\tau$ and $\sum_y p^*(y\mid x)=1$, the cross entropy equals $-\sum_y p^*(y\mid x)\,\ell_y+\log\sum_{y}e^{\ell_y}$. The first term is linear in $q$ and the second is a log sum exp of linear functions of $q$, which is convex. Its gradient is $\tfrac1\tau\sum_y p_q(y\mid x)\,g_y$, which gives the stated gradient, and a convex differentiable function is minimized exactly where its gradient vanishes, which gives the condition of Section~3.2 of the main paper. Under the constraint $\|q\|=1$, a stationary point satisfies $(I-qq^\top)\nabla_q\mathcal{L}=0$, so the condition holds for the component tangent to the sphere.

\paragraph{Loss~(1).}
The cross entropy is linear in its first argument, so $\mathrm{CE}(\bar p_\lambda,p_q)=(1-\lambda)\,\mathrm{CE}(\delta_{y_x},p_q)+\lambda\,\mathrm{CE}(p_{\mathrm{new}},p_q)$, and $\mathrm{CE}(p_{\mathrm{new}},p_q)=\mathrm{KL}(p_{\mathrm{new}}\|p_q)+H(p_{\mathrm{new}})$, where the entropy $H(p_{\mathrm{new}})$ does not depend on $q$.

\paragraph{Stationarity of the linear head.}
Let $u=Wz+b$ and $q=\mathrm{unit}(u)$, so that $\partial q/\partial u=J=(I-qq^\top)/\|u\|$. By the chain rule, $\nabla_W\mathbb{E}_x\mathcal{L}=\tfrac1\tau\,\mathbb{E}_x\big[J\,(\mathbb{E}_{p_q}[g]-\mathbb{E}_{\bar p}[g])\,z^\top\big]$. At a stationary point this matrix vanishes: the residual of the optimality condition, projected onto the tangent space and rescaled by $1/\|u\|$, is uncorrelated with every coordinate of $z$.

\subsection{View barycenters}

\paragraph{Closed form~(2).}
For a unit vector $u$, $\|u-q_v\|^2=2-2\langle u,q_v\rangle$, hence $\sum_v c_v\|u-q_v\|^2=2-2\langle u,\bar q\rangle$. By the Cauchy Schwarz inequality, $\langle u,\bar q\rangle\le\|\bar q\|$ with equality only at $u=\mathrm{unit}(\bar q)$ when $\bar q\ne0$. The score identity $\langle \bar q,g_y\rangle=\sum_v c_v s_v(x,y)$ is linearity of the inner product, and dividing by $\|\bar q\|>0$ does not change the ranking.

\paragraph{Proof of Proposition~1.}
Since $c_v\ge0$ and the scores of the barycenter are $\langle \bar q,g\rangle/\|\bar q\|$, its gaps are $\Delta_B/\|\bar q\|$, which have the same signs and order as $\Delta_B$. The rank of $y^+$ is one plus the number of irrelevant items with $\Delta_B(n)\le0$, so $y^+$ is among the top $k$ exactly when $\Delta_{B,(k)}>0$.

(i) For every $n$, $\Delta_B(n)=\sum_v c_v\Delta_v(n)\ge\sum_v c_v m_v$, so $m_B\ge\sum_v c_v m_v$ and $D\ge0$. Evaluating at $\hat n$, where $\Delta_0(\hat n)=m_0$, gives $m_B\le\Delta_B(\hat n)=\sum_v c_v\big(m_v+\gamma_v(\hat n)\big)=\sum_v c_v m_v+\sum_{v\ne0}c_v\gamma_v(\hat n)$, which is the upper bound. Let $n^\star$ attain $m_B$. Then $D=\sum_v c_v\big(\Delta_v(n^\star)-m_v\big)$ is a sum of nonnegative terms, so $D=0$ implies $\Delta_v(n^\star)=m_v$ for every $v$ with $c_v>0$, that is, $n^\star$ is the hardest item of all these views. Conversely, if one item $n_0$ is the hardest for all of them, $m_B\le\Delta_B(n_0)=\sum_v c_v m_v$ and therefore $D=0$.

(ii) Let $E_v=\{n\in\mathcal{N}:\Delta_v(n)<\Delta_{v,(k_v)}\}$. By definition of the order statistic, $|E_v|\le k_v-1$, so $|\bigcup_v E_v|\le\sum_v(k_v-1)=k-1$. Let $S$ be a set of $k$ items with the smallest values of $\Delta_B$. Since $|S|=k$, some $n'\in S$ lies outside every $E_v$, so $\Delta_v(n')\ge\Delta_{v,(k_v)}$ for all $v$ and, as $c_v\ge0$, $\Delta_B(n')\ge\sum_v c_v\Delta_{v,(k_v)}$. Finally $\Delta_{B,(k)}=\max_{n\in S}\Delta_B(n)\ge\Delta_B(n')$. \qed

\paragraph{Effect of an additional view.}
With $V$ views of equal weight, adding view $j$ gives $\Delta^{V+1}(n)=(V\Delta^V(n)+\Delta_j(n))/(V+1)$ for every $n$. Since $\Delta^{V+1}(n)\ge(V\Delta^V(n)+\Delta_{j,(1)})/(V+1)$, the first fact on order statistics gives $\Delta^{V+1}_{(k)}-\Delta^V_{(k)}\ge(\Delta_{j,(1)}-\Delta^V_{(k)})/(V+1)$. With $S_k$ the $k$ items with the smallest $\Delta^V$, the second fact gives $\Delta^{V+1}_{(k)}\le\max_{n\in S_k}(V\Delta^V(n)+\Delta_j(n))/(V+1)\le(V\Delta^V_{(k)}+\max_{n\in S_k}\Delta_j(n))/(V+1)$. Both changes are of order $1/(V+1)$.

\paragraph{Minimum variance weights.}
Under the noise model, $\sum_v w_v(s_v-b_v)/a_v=r+\sum_v w_v\varepsilon_v/a_v$ whenever $\sum_v w_v=1$, with noise variance $w^\top\Sigma w$. For positive definite $\Sigma$, minimizing $w^\top\Sigma w$ subject to $\mathbf 1^\top w=1$ with a Lagrange multiplier gives $\Sigma w=\mu\mathbf 1$, hence $w^*=\Sigma^{-1}\mathbf 1/(\mathbf 1^\top\Sigma^{-1}\mathbf 1)$ with variance $V^*=1/(\mathbf 1^\top\Sigma^{-1}\mathbf 1)\le\Sigma_{vv}$ for every $v$. The fused score equals $\langle\sum_v(w_v/a_v)q_v,g_y\rangle$ minus a term constant in $y$, so the barycenter weights are $c_v\propto w^*_v/a_v$. For two views, $\Sigma=\bigl(\begin{smallmatrix}\sigma_0^2&\rho\sigma_0\sigma_a\\\rho\sigma_0\sigma_a&\sigma_a^2\end{smallmatrix}\bigr)$ and $\mathbf 1^\top\Sigma^{-1}\mathbf 1=(\sigma_0^2+\sigma_a^2-2\rho\sigma_0\sigma_a)/(\sigma_0^2\sigma_a^2(1-\rho^2))$, which gives $\Gamma=1-V^*/\sigma_0^2$. With weight $t\in[0,1]$ on the extra view, the variance $\nu(t)=(1-t)^2\sigma_0^2+t^2\sigma_a^2+2t(1-t)\rho\sigma_0\sigma_a$ is a convex quadratic with $\nu'(0)=2\sigma_0(\rho\sigma_a-\sigma_0)$, so a small positive weight lowers the variance of view $0$ exactly when $\rho<\sigma_0/\sigma_a$.

\subsection{Inheriting the consensus}

\paragraph{Margin preservation.}
By the Cauchy Schwarz inequality, $|\langle q_\theta-T,g\rangle|\le\|q_\theta-T\|\,\|g\|\le\delta$ for every unit $g$. Let $y^+$ be the item that the teacher ranks first and $m_T$ its margin. For every other item $y$, $s_{q_\theta}(y^+)-s_{q_\theta}(y)\ge s_T(y^+)-s_T(y)-2\delta\ge m_T-2\delta$, which is positive when $\delta<m_T/2$.

\paragraph{Barycentric regression.}
For every vector $u$, expanding the squares gives $\sum_v c_v\|u-q_v\|^2=\|u-\bar q\|^2+\sum_v c_v\|q_v-\bar q\|^2$, and with unit views $\sum_v c_v\|q_v-\bar q\|^2=1-\|\bar q\|^2$. For a unit $q$, $\|q-T\|^2=2-2\langle q,\bar q\rangle/\|\bar q\|$ and $\|q-\bar q\|^2=1-2\langle q,\bar q\rangle+\|\bar q\|^2$. Eliminating $\langle q,\bar q\rangle$ gives $\|q-T\|^2=\big(\|q-\bar q\|^2-(1-\|\bar q\|)^2\big)/\|\bar q\|$, and substituting $\|q-\bar q\|^2=\sum_v c_v\|q-q_v\|^2-(1-\|\bar q\|^2)$ gives the stated identity, since $(1-\|\bar q\|^2)+(1-\|\bar q\|)^2=2(1-\|\bar q\|)$.

\paragraph{Projection onto the student family.}
Let $\mathcal{H}$ be the Hilbert space of square integrable maps $x\mapsto\mathbb{R}^d$ under the training distribution, with $\langle u_1,u_2\rangle=\mathbb{E}_x\langle u_1(x),u_2(x)\rangle$. For $q=q_0+l$ with $l$ in a closed subspace $\mathcal{L}$, the first term of~(4) equals $\|l-G\|^2$. By the projection theorem, $\mathcal{L}$ contains a unique minimizer $\Pi_{\mathcal{L}}G$, and $G-\Pi_{\mathcal{L}}G$ is orthogonal to $\mathcal{L}$. By Pythagoras, $\|G\|^2=\|\Pi_{\mathcal{L}}G\|^2+\|G-\Pi_{\mathcal{L}}G\|^2$, so $\eta=\|\Pi_{\mathcal{L}}G\|^2/\|G\|^2=1-\|G-\Pi_{\mathcal{L}}G\|^2/\|G\|^2$, the coefficient of determination of the least squares fit of $G$ within $\mathcal{L}$. This treats the family as affine and ignores the normalization of the query; for LoRA it applies to the first order expansion $q_0+J(\theta-\theta_0)$.

\section{Stage 1 Sweeps}
\label{sec:supp_sweeps}
Tables~\ref{tab:supp_lambda} and~\ref{tab:supp_beta} give the full sweeps of the mixing weight $\lambda$ of the stage 1 target and of the weight $\beta$ of $q_0$, summarized in Section~4.4 of the main paper.

\begin{table*}[t]
\centering
\small
\caption{Mixing weight $\lambda$ of the stage 1 target (stage 1 query $q_0$; mean R@1 over the five sets, image $\to$ text / text $\to$ image, \%). $\lambda=0$ uses the label only and $\lambda=1$ the ranking of the new model only; $\lambda^*$ is the minimum variance weight estimated on COCO DEV.}
\label{tab:supp_lambda}
\begin{tabular}{lcccccc}
\toprule
Pair & $\lambda=0$ & $0.25$ & $0.5$ & $0.75$ & $1$ & $\lambda^*$ \\
\midrule
B/32 $\to$ L/14 & 48.9 / 33.2 & 50.8 / 33.9 & 52.4 / 34.4 & 53.1 / 34.2 & 51.7 / 32.2 & 53.0 / 34.0 (0.82) \\
B/32 $\to$ B/16 & 47.1 / 32.9 & 49.7 / 33.7 & 51.2 / 33.9 & 51.9 / 33.4 & 50.6 / 30.9 & 51.9 / 33.1 (0.81) \\
B/32 $\to$ SO400M & 56.0 / 37.2 & 56.9 / 37.8 & 57.6 / 38.5 & 57.8 / 39.0 & 55.9 / 38.8 & 57.5 / 39.1 (0.84) \\
B/32 $\to$ MetaCLIP L/14 & 51.9 / 36.1 & 52.5 / 36.6 & 53.3 / 37.1 & 53.0 / 37.0 & 51.6 / 36.5 & 53.0 / 37.1 (0.83) \\
MetaCLIP B/32 $\to$ SO400M & 62.6 / 38.7 & 63.4 / 39.3 & 63.8 / 40.0 & 64.0 / 40.4 & 62.9 / 40.0 & 63.9 / 40.4 (0.84) \\
\bottomrule
\end{tabular}
\end{table*}

\begin{table*}[t]
\centering
\small
\caption{Weight $\beta$ of the stage 1 head in $q_0=\mathrm{unit}\big(\beta\,f(z)+(1-\beta)\,\mathrm{unit}(P^\top z)\big)$ (mean R@1 over the five sets, image $\to$ text / text $\to$ image, \%). $\beta=0$ is the Procrustes map alone, $\beta=1$ the head alone; the deployed value is $\beta=0.75$.}
\label{tab:supp_beta}
\begin{tabular}{lccccc}
\toprule
Pair & $\beta=0$ & $0.25$ & $0.5$ & $0.75$ & $1$ \\
\midrule
B/32 $\to$ L/14 & 48.8 / 30.3 & 52.1 / 33.1 & 53.2 / 34.5 & 52.4 / 34.4 & 50.2 / 32.9 \\
B/32 $\to$ B/16 & 48.6 / 28.8 & 51.0 / 32.0 & 52.1 / 33.7 & 51.2 / 33.9 & 49.6 / 33.1 \\
B/32 $\to$ SO400M & 46.8 / 32.8 & 55.3 / 36.7 & 59.0 / 38.6 & 57.6 / 38.5 & 52.9 / 36.8 \\
B/32 $\to$ MetaCLIP L/14 & 50.9 / 35.0 & 53.9 / 36.8 & 54.5 / 37.5 & 53.3 / 37.1 & 50.9 / 35.5 \\
MetaCLIP B/32 $\to$ SO400M & 56.3 / 37.2 & 61.4 / 39.4 & 63.7 / 40.1 & 63.8 / 40.0 & 62.3 / 38.7 \\
\bottomrule
\end{tabular}
\end{table*}

\section{Empirical Analysis of the Theory}
\label{sec:supp_theory}
\begin{table*}[t]
\centering
\small
\caption{Consensus correction on the test sets, extending Table~2 of the main paper (image $\to$ text, one designated positive caption per image, other positives removed from the negatives; mean over the five sets, \%). Rescued: the barycenter ranks the positive first while the new view does not. Needing $D$: rescued although the weighted average of the view margins is not positive. All views wrong: rescued although every view ranks the positive lower. Shared: all three views have the same hardest irrelevant item.}
\label{tab:supp_consensus}
\begin{tabular}{lrrrrrrr}
\toprule
Pair & New view & Barycenter & Ceiling & Rescued & Needing $D$ & All views wrong & Shared \\
\midrule
B/32 $\to$ L/14 & 23.0 & 27.7 & 38.0 & 8.0 & 4.7 & 1.2 & 8.7 \\
B/32 $\to$ B/16 & 22.9 & 27.5 & 37.1 & 7.6 & 4.4 & 1.1 & 9.9 \\
B/32 $\to$ SO400M & 24.3 & 29.5 & 44.3 & 9.4 & 5.7 & 1.7 & 7.0 \\
B/32 $\to$ MetaCLIP L/14 & 23.0 & 28.3 & 40.8 & 9.0 & 5.2 & 1.4 & 8.0 \\
MetaCLIP B/32 $\to$ SO400M & 32.3 & 33.6 & 40.6 & 6.0 & 3.4 & 1.0 & 11.6 \\
\bottomrule
\end{tabular}
\end{table*}

Proposition~1 makes two exact claims that can be checked query by query: the barycentric margin is never below the weighted average of the view margins, and it never exceeds the ceiling given by the spread of the other views at the hardest item of view~0. Both held for every test query of every pair. Table~\ref{tab:supp_consensus} quantifies the correction. Under this protocol the barycenter of the three views raises R@1 over the new view by 1.3 to 5.3 points and rescues 6.0\% to 9.4\% of the queries. Close to 60\% of the rescued queries, 3.4\% to 5.7\% of all queries, have a nonpositive average view margin and are rescued only through the dispersion gain $D$, and 1.0\% to 1.7\% of all queries are ranked correctly by the barycenter although every view ranks them wrongly. All three views share their hardest irrelevant item on 7.0\% to 11.6\% of the queries; when such a query is ranked wrongly by every view, no weighting rescues it. Splitting the queries by this condition confirms the proposition: on the queries with a shared hardest item the measured $D$ is zero up to numerical precision (at most $7\times10^{-8}$) and none of those that every view ranks wrongly is rescued, whereas $D$ is positive on more than 99.9\% of the remaining queries, where the barycenter corrects 10.0\% to 13.4\% of the errors of the new view, against 4.0\% to 4.5\% on the queries with a shared hardest item.

The noise model of Section~3.3 of the main paper is estimated on CC3M DEV only. In a preregistered test over 40 combinations of a model pair and a candidate view, its predicted gain $\Gamma$ ranks the views by their measured test gain (Spearman 0.38, one sided permutation $p=0.007$; 0.61 on average within a pair), whereas the correlation $\rho$ alone does not (0.08). The same model overestimates the absolute size of the gains by a factor of three to four, because it treats the noise of different irrelevant items as independent while the errors of the views concentrate on a few shared hard items, the situation in which Proposition~1 predicts no dispersion gain. Training the student on the teacher with the derived weights $c\propto w^*/a$ gives image $\to$ text R@1 on par with a fixed rule without estimation, the nested midpoint $\mathrm{unit}\big(q_{\mathrm{new}}+\mathrm{unit}(q_{\mathrm{old}}+q_{\mathrm{aux}})\big)$ with vector weights of roughly $\tfrac12,\tfrac14,\tfrac14$: over the 25 combinations of pair and set, the mean difference is $-0.05$ points (mean over three seeds), positive in 12.

\section{Full Ablation}
\label{sec:supp_ablation}

\begin{table*}[t]
\centering
\footnotesize
\setlength{\tabcolsep}{4pt}
\caption{Full ablation, extending Table~3 of the main paper. Contribution of each component (image $\to$ text R@1, mean over the five datasets, \%; students with seed 0). Stage 1: the Procrustes map alone ($\beta=0$), the head alone ($\beta=1$) and $q_0$. Single views and teachers are queries in the old space; the teacher of two views drops the auxiliary encoder. Students: trained on the teacher of two views without the auxiliary term, on the teacher of three views without the auxiliary term ($\mu=0$), and \ours.}
\label{tab:supp_ablation}
\begin{tabular}{lrrrrrrrrrr}
\toprule
& \multicolumn{3}{c}{Stage 1} & \multicolumn{2}{c}{Single view} & \multicolumn{2}{c}{Teacher} & \multicolumn{3}{c}{Student} \\
\cmidrule(lr){2-4} \cmidrule(lr){5-6} \cmidrule(lr){7-8} \cmidrule(lr){9-11}
Pair & Procrustes & Head & $q_0$ & Old & H/14 & 2 views & 3 views & 2 views & 3 views, $\mu=0$ & \ours \\
\midrule
B/32 $\to$ L/14 & 48.8 & 50.2 & 52.4 & 49.1 & 50.2 & 56.0 & 56.6 & 54.8 & 54.8 & 55.8 \\
B/32 $\to$ B/16 & 48.6 & 49.6 & 51.2 & 49.1 & 50.2 & 54.3 & 55.8 & 52.7 & 52.6 & 53.2 \\
B/32 $\to$ SO400M & 46.8 & 52.9 & 57.6 & 49.1 & 50.2 & 58.8 & 59.7 & 58.8 & 58.6 & 58.8 \\
B/32 $\to$ MetaCLIP L/14 & 50.9 & 50.9 & 53.3 & 49.1 & 50.2 & 56.8 & 57.4 & 55.5 & 55.2 & 55.8 \\
MetaCLIP B/32 $\to$ SO400M & 56.3 & 62.3 & 63.8 & 54.0 & 54.5 & 64.5 & 64.9 & 64.5 & 64.1 & 64.2 \\
\bottomrule
\end{tabular}
\end{table*}

The full ablation adds the head alone, the single views, and the teachers and students of two views to the ablation of the main paper (Table~\ref{tab:supp_ablation}). Each part of the stage 1 query contributes: the Procrustes map alone is 2.5 to 10.9 points below $q_0$ and the head alone 1.5 to 4.7 points. The two estimates of relevance are both needed (Table~\ref{tab:supp_lambda}): the label alone ($\lambda=0$) is 1.2 to 4.1 points below $\lambda=\tfrac12$ and the ranking of the new model alone ($\lambda=1$) 0.6 to 1.7 points below.

The barycenter of three views is the strongest target, and the auxiliary term carries the third view into the student. The teacher of three views improves on $q_0$ by 1.1 to 4.5 points and on the teacher of two views by 0.4 to 1.4 points, and every teacher exceeds each of its views. Without the auxiliary term, the student of three views ($\mu=0$) is within 0.5 points of the student of two views, although its teacher is better, and adding it gives 0.1 to 1.0 points. On MetaCLIP B/32 $\to$ SO400M, where the teacher gains least, the student of two views is 0.3 points ahead of \ours.

\section{Training Cost}
\label{sec:supp_cost}

\begin{table*}[t]
\centering
\small
\caption{Training cost in A100 hours (bf16, preprocessing and tokenization excluded, one configuration and one seed), counting the encoding of the training data with every encoder a method needs and the optimization, and the cost of a full backfill with the new model.}
\label{tab:supp_cost}
\resizebox{\textwidth}{!}{%
\begin{tabular}{lrrrrr}
\toprule
Method & B/32 $\to$ L/14 & B/32 $\to$ B/16 & B/32 $\to$ SO400M & B/32 $\to$ MetaCLIP L/14 & MetaCLIP B/32 $\to$ SO400M \\
\midrule
XBT~\cite{jang2025xbt}, paper scale & 10.03 & 4.18 & 18.94 & 10.03 & 18.93 \\
\ours (ours) & 0.44 & 0.22 & 0.71 & 0.44 & 0.71 \\
\midrule
Full backfill, per 1M images & 0.39 & 0.10 & 0.51 & 0.39 & 0.51 \\
Full backfill, per 1B images & 388 & 101 & 506 & 389 & 506 \\
Gallery where backfill costs as much as \ours & 1.1M & 2.2M & 1.4M & 1.1M & 1.4M \\
\bottomrule
\end{tabular}}
\end{table*}

\ours trains 19 to 27 times faster than XBT, the other method that trains the new encoder for compatibility across modalities (Table~\ref{tab:supp_cost}). \ours needs 0.22 to 0.71 A100 hours, spent on encoding its training data with the old, new and auxiliary encoders and on the student, while XBT at the scale of its paper needs 4.2 to 18.9 hours. Query side maps fit in minutes on cached features.

Training \ours costs as much as reencoding 1.1M to 2.2M images, and this cost does not grow with the gallery. A full backfill with the new model costs 101 to 506 A100 hours per billion images.

\section{Update Gain}
\label{sec:supp_gain}

\begin{table*}[t]
\centering
\small
\setlength{\tabcolsep}{4pt}
\caption{Update gain~\cite{shen2020bct} (\%), the share of the R@1 gain of reencoding that a method obtains without reencoding, over all pairs and datasets.}
\label{tab:supp_gain}
\begin{tabular}{lcccc}
\toprule
Method & Encoders & Image $\to$ text & Text $\to$ image & All \\
\midrule
Drift-Adapter~\cite{vejendla2025drift} & 1 & 14.0 & 0.4 & 6.8 \\
LST~\cite{maiorca2023lst} & 1 & 4.2 & $-7.5$ & $-2.0$ \\
$\lambda$-orth~\cite{ricci2025lorth} & 1 & $-17.6$ & 3.0 & $-6.7$ \\
XBT~\cite{jang2025xbt} & 1 & $-7.0$ & 6.2 & 0.0 \\
SLERP~\cite{ricci2026slerp} & 2 & 54.4 & 19.9 & 36.1 \\
\rowcolor{ourblue} \ours (ours) & 1 & \textbf{68.4} & \textbf{43.3} & \textbf{55.0} \\
\bottomrule
\end{tabular}
\end{table*}

\ours obtains more than half of the gain of reencoding the gallery without reencoding it (Table~\ref{tab:supp_gain}). We report the update gain of BCT~\cite{shen2020bct}, the improvement of a method over Old $\to$ old divided by the improvement of New $\to$ new, on the mean R@1 over the five pairs and five datasets, so that 0\% keeps the old model and 100\% matches reencoding. \ours reaches 55.0\%, with 68.4\% from image to text and 43.3\% from text to image.

Methods with one encoder obtain almost none of this gain, and SLERP obtains less with two encoders. Query side maps and XBT range from $-6.7\%$ to $6.8\%$, as expected from methods that fit the old query, and SLERP reaches 36.1\% while it runs the old encoder for every query. The gap is largest from text to image, where \ours obtains 43.3\% and SLERP 19.9\%.

\end{document}









%% file: main.bbl
\begin{thebibliography}{51}
\providecommand{\natexlab}[1]{#1}
\providecommand{\url}[1]{\texttt{#1}}
\expandafter\ifx\csname urlstyle\endcsname\relax
  \providecommand{\doi}[1]{doi: #1}\else
  \providecommand{\doi}{doi: \begingroup \urlstyle{rm}\Url}\fi

\bibitem[Agrawal et~al.(2019)Agrawal, Desai, Wang, Chen, Jain, Johnson, Batra,
  Parikh, Lee, and Anderson]{agrawal2019nocaps}
Harsh Agrawal, Karan Desai, Yufei Wang, Xinlei Chen, Rishabh Jain, Mark
  Johnson, Dhruv Batra, Devi Parikh, Stefan Lee, and Peter Anderson.
\newblock Nocaps: Novel object captioning at scale.
\newblock In \emph{2019 IEEE/CVF International Conference on Computer Vision
  (ICCV)}, pages 8947--8956. IEEE, 2019.

\bibitem[Bates and Granger(1969)]{bates1969combination}
John~M Bates and Clive~WJ Granger.
\newblock The combination of forecasts.
\newblock \emph{Journal of the operational research society}, 20\penalty0
  (4):\penalty0 451--468, 1969.

\bibitem[Biondi et~al.(2023)Biondi, Pernici, Bruni, and
  Del~Bimbo]{biondi2023cores}
Niccolò Biondi, Federico Pernici, Matteo Bruni, and Alberto Del~Bimbo.
\newblock Cores: Compatible representations via stationarity.
\newblock \emph{IEEE Transactions on Pattern Analysis and Machine
  Intelligence}, 45\penalty0 (8):\penalty0 9567--9582, 2023.

\bibitem[Biondi et~al.(2024)Biondi, Pernici, Ricci, and
  Del~Bimbo]{biondi2024stationary}
Niccol{\`o} Biondi, Federico Pernici, Simone Ricci, and Alberto Del~Bimbo.
\newblock Stationary representations: Optimally approximating compatibility and
  implications for improved model replacements.
\newblock In \emph{2024 IEEE/CVF Conference on Computer Vision and Pattern
  Recognition (CVPR)}, pages 28793--28804. IEEE, 2024.

\bibitem[Byun et~al.(2026)Byun, Kwon, Hsu, Karumuri, Zhang, Yang, and
  Modolo]{byun2026mkp}
Jaeseok Byun, Gukyeong Kwon, Han-Kai Hsu, Meher~Gitika Karumuri, Zhikang Zhang,
  Hao Yang, and Davide Modolo.
\newblock Multi-modal knowledge preserving adapter for embedding backward
  compatibility.
\newblock In \emph{European Conference on Computer Vision}, pages 473--490.
  Springer, 2026.

\bibitem[Cao et~al.(2007)Cao, Qin, Liu, Tsai, and Li]{cao2007listnet}
Zhe Cao, Tao Qin, Tie-Yan Liu, Ming-Feng Tsai, and Hang Li.
\newblock Learning to rank: from pairwise approach to listwise approach.
\newblock In \emph{Proceedings of the 24th International Conference on Machine
  Learning}, page 129–136, New York, NY, USA, 2007. Association for Computing
  Machinery.

\bibitem[Cherti et~al.(2023)Cherti, Beaumont, Wightman, Wortsman, Ilharco,
  Gordon, Schuhmann, Schmidt, and Jitsev]{cherti2023openclip}
Mehdi Cherti, Romain Beaumont, Ross Wightman, Mitchell Wortsman, Gabriel
  Ilharco, Cade Gordon, Christoph Schuhmann, Ludwig Schmidt, and Jenia Jitsev.
\newblock Reproducible scaling laws for contrastive language-image learning.
\newblock In \emph{2023 IEEE/CVF Conference on Computer Vision and Pattern
  Recognition (CVPR)}, pages 2818--2829. IEEE, 2023.

\bibitem[Duggal et~al.(2021)Duggal, Zhou, Yang, Xiong, Xia, Tu, and
  Soatto]{duggal2021compat}
Rahul Duggal, Hao Zhou, Shuo Yang, Yuanjun Xiong, Wei Xia, Zhuowen Tu, and
  Stefano Soatto.
\newblock Compatibility-aware heterogeneous visual search.
\newblock In \emph{2021 IEEE/CVF Conference on Computer Vision and Pattern
  Recognition (CVPR)}, pages 10718--10727, 2021.

\bibitem[Goswami et~al.(2025)Goswami, Wang, Twardowski, and van~de
  Weijer]{goswami2025qdc}
Dipam Goswami, Liying Wang, Bart{\l}omiej Twardowski, and Joost van~de Weijer.
\newblock Query drift compensation: Enabling compatibility in continual
  learning of retrieval embedding models.
\newblock \emph{arXiv preprint arXiv:2506.00037}, 2025.

\bibitem[Hinton et~al.(2015)Hinton, Vinyals, and Dean]{hinton2015distilling}
Geoffrey Hinton, Oriol Vinyals, and Jeff Dean.
\newblock Distilling the knowledge in a neural network.
\newblock \emph{arXiv preprint arXiv:1503.02531}, 2015.

\bibitem[Hu et~al.(2021)Hu, Shen, Wallis, Allen-Zhu, Li, Wang, Wang, and
  Chen]{hu2022lora}
Edward~J Hu, Yelong Shen, Phillip Wallis, Zeyuan Allen-Zhu, Yuanzhi Li, Shean
  Wang, Lu Wang, and Weizhu Chen.
\newblock Lora: Low-rank adaptation of large language models.
\newblock \emph{arXiv preprint arXiv:2106.09685}, 2021.

\bibitem[Huh et~al.(2024)Huh, Cheung, Wang, and Isola]{huh2024platonic}
Minyoung Huh, Brian Cheung, Tongzhou Wang, and Phillip Isola.
\newblock The platonic representation hypothesis, 2024.

\bibitem[Jaeckle et~al.(2023)Jaeckle, Faghri, Farhadi, Tuzel, and
  Pouransari]{jaeckle2023fastfill}
Florian Jaeckle, Fartash Faghri, Ali Farhadi, Oncel Tuzel, and Hadi Pouransari.
\newblock Fastfill: Efficient compatible model update.
\newblock \emph{arXiv preprint arXiv:2303.04766}, 2023.

\bibitem[Jang and Lim(2025)]{jang2025xbt}
Young~Kyun Jang and Ser-nam Lim.
\newblock Towards cross-modal backward-compatible representation learning for
  vision-language models.
\newblock In \emph{2025 IEEE/CVF International Conference on Computer Vision
  (ICCV)}, pages 1783--1792. IEEE, 2025.

\bibitem[Jia et~al.(2021)Jia, Yang, Xia, Chen, Parekh, Pham, Le, Sung, Li, and
  Duerig]{jia2021align}
Chao Jia, Yinfei Yang, Ye Xia, Yi-Ting Chen, Zarana Parekh, Hieu Pham, Quoc Le,
  Yun-Hsuan Sung, Zhen Li, and Tom Duerig.
\newblock Scaling up visual and vision-language representation learning with
  noisy text supervision.
\newblock In \emph{International conference on machine learning}, pages
  4904--4916. PMLR, 2021.

\bibitem[Johnson et~al.(2019)Johnson, Douze, and J{\'e}gou]{johnson2019faiss}
Jeff Johnson, Matthijs Douze, and Herv{\'e} J{\'e}gou.
\newblock Billion-scale similarity search with gpus.
\newblock \emph{IEEE transactions on big data}, 7\penalty0 (3):\penalty0
  535--547, 2019.

\bibitem[Krogh and Vedelsby(1994)]{krogh1995ensembles}
Anders Krogh and Jesper Vedelsby.
\newblock Neural network ensembles, cross validation, and active learning.
\newblock \emph{Advances in neural information processing systems}, 7, 1994.

\bibitem[Li et~al.(2021)Li, Selvaraju, Gotmare, Joty, Xiong, and
  Hoi]{li2021albef}
Junnan Li, Ramprasaath Selvaraju, Akhilesh Gotmare, Shafiq Joty, Caiming Xiong,
  and Steven Chu~Hong Hoi.
\newblock Align before fuse: Vision and language representation learning with
  momentum distillation.
\newblock \emph{Advances in neural information processing systems},
  34:\penalty0 9694--9705, 2021.

\bibitem[Lin et~al.(2014)Lin, Maire, Belongie, Hays, Perona, Ramanan,
  Doll{\'a}r, and Zitnick]{lin2014coco}
Tsung-Yi Lin, Michael Maire, Serge Belongie, James Hays, Pietro Perona, Deva
  Ramanan, Piotr Doll{\'a}r, and C~Lawrence Zitnick.
\newblock Microsoft coco: Common objects in context.
\newblock In \emph{European conference on computer vision}, pages 740--755.
  Springer, 2014.

\bibitem[Maiorca et~al.(2023)Maiorca, Moschella, Norelli, Fumero, Locatello,
  and Rodol{\`a}]{maiorca2023lst}
Valentino Maiorca, Luca Moschella, Antonio Norelli, Marco Fumero, Francesco
  Locatello, and Emanuele Rodol{\`a}.
\newblock Latent space translation via semantic alignment.
\newblock \emph{Advances in Neural Information Processing Systems},
  36:\penalty0 55394--55414, 2023.

\bibitem[Meng et~al.(2021)Meng, Zhang, Xu, and Zhou]{meng2021lce}
Qiang Meng, Chixiang Zhang, Xiaoqiang Xu, and Feng Zhou.
\newblock Learning compatible embeddings.
\newblock In \emph{2021 IEEE/CVF International Conference on Computer Vision
  (ICCV)}, pages 9919--9928. IEEE, 2021.

\bibitem[Moschella et~al.(2022)Moschella, Maiorca, Fumero, Norelli, Locatello,
  and Rodol{\`a}]{moschella2023relative}
Luca Moschella, Valentino Maiorca, Marco Fumero, Antonio Norelli, Francesco
  Locatello, and Emanuele Rodol{\`a}.
\newblock Relative representations enable zero-shot latent space communication.
\newblock \emph{arXiv preprint arXiv:2209.15430}, 2022.

\bibitem[Pan et~al.(2023)Pan, Xu, Yang, He, Jiang, Guo, Qian, Zhang, Cheng,
  Yang, et~al.]{pan2023advbct}
Tan Pan, Furong Xu, Xudong Yang, Sifeng He, Chen Jiang, Qingpei Guo, Feng Qian,
  Xiaobo Zhang, Yuan Cheng, Lei Yang, et~al.
\newblock Boundary-aware backward-compatible representation via adversarial
  learning in image retrieval.
\newblock In \emph{2023 IEEE/CVF Conference on Computer Vision and Pattern
  Recognition (CVPR)}, pages 15201--15210. IEEE, 2023.

\bibitem[Parekh et~al.(2021)Parekh, Baldridge, Cer, Waters, and
  Yang]{parekh2021cxc}
Zarana Parekh, Jason Baldridge, Daniel Cer, Austin Waters, and Yinfei Yang.
\newblock Crisscrossed captions: Extended intramodal and intermodal semantic
  similarity judgments for ms-coco.
\newblock In \emph{Proceedings of the 16th Conference of the European Chapter
  of the Association for Computational Linguistics: Main Volume}, pages
  2855--2870, 2021.

\bibitem[Park et~al.(2019)Park, Kim, Lu, and Cho]{park2019rkd}
Wonpyo Park, Dongju Kim, Yan Lu, and Minsu Cho.
\newblock Relational knowledge distillation.
\newblock In \emph{2019 IEEE/CVF Conference on Computer Vision and Pattern
  Recognition (CVPR)}, pages 3962--3971. IEEE, 2019.

\bibitem[Radford et~al.(2021)Radford, Kim, Hallacy, Ramesh, Goh, Agarwal,
  Sastry, Askell, Mishkin, Clark, et~al.]{radford2021clip}
Alec Radford, Jong~Wook Kim, Chris Hallacy, Aditya Ramesh, Gabriel Goh,
  Sandhini Agarwal, Girish Sastry, Amanda Askell, Pamela Mishkin, Jack Clark,
  et~al.
\newblock Learning transferable visual models from natural language
  supervision.
\newblock In \emph{International conference on machine learning}, pages
  8748--8763. PmLR, 2021.

\bibitem[Ramanujan et~al.(2022)Ramanujan, Vasu, Farhadi, Tuzel, and
  Pouransari]{ramanujan2022fct}
Vivek Ramanujan, Pavan Kumar~Anasosalu Vasu, Ali Farhadi, Oncel Tuzel, and Hadi
  Pouransari.
\newblock Forward compatible training for large-scale embedding retrieval
  systems.
\newblock In \emph{2022 IEEE/CVF Conference on Computer Vision and Pattern
  Recognition (CVPR)}, pages 19364--19373. IEEE, 2022.

\bibitem[Ricci et~al.(2024)Ricci, Biondi, Pernici, and
  Del~Bimbo]{ricci2024orthogonal}
Simone Ricci, Niccol{\`o} Biondi, Federico Pernici, and Alberto Del~Bimbo.
\newblock Backward-compatible aligned representations via an orthogonal
  transformation layer.
\newblock In \emph{European Conference on Computer Vision}, pages 451--464.
  Springer, 2024.

\bibitem[Ricci et~al.(2025)Ricci, Biondi, Pernici, Patras, and
  Del~Bimbo]{ricci2025lorth}
Simone Ricci, Niccol{\`o} Biondi, Federico Pernici, Ioannis Patras, and Alberto
  Del~Bimbo.
\newblock {$\boldsymbol{\lambda}$}-orthogonality regularization for compatible
  representation learning.
\newblock \emph{Advances in Neural Information Processing Systems},
  38:\penalty0 29036--29063, 2025.

\bibitem[Ricci et~al.(2026)Ricci, Biondi, and Pernici]{ricci2026slerp}
Simone Ricci, Niccol{\`o} Biondi, and Federico Pernici.
\newblock Spherical interpolation for backward-compatible multimodal
  representations.
\newblock \emph{arXiv preprint arXiv:2609.39836}, 2026.

\bibitem[Romero et~al.(2015)Romero, Ballas, Kahou, Chassang, Gatta, and
  Bengio]{romero2015fitnets}
Adriana Romero, Nicolas Ballas, Samira~Ebrahimi Kahou, Antoine Chassang, Carlo
  Gatta, and Yoshua Bengio.
\newblock Fitnets: Hints for thin deep nets, 2015.

\bibitem[Sch{\"o}nemann(1966)]{schonemann1966procrustes}
Peter~H Sch{\"o}nemann.
\newblock A generalized solution of the orthogonal procrustes problem.
\newblock \emph{Psychometrika}, 31\penalty0 (1):\penalty0 1--10, 1966.

\bibitem[Schuhmann et~al.(2022)Schuhmann, Beaumont, Vencu, Gordon, Wightman,
  Cherti, Coombes, Katta, Mullis, Wortsman, Schramowski, Kundurthy, Crowson,
  Schmidt, Kaczmarczyk, and Jitsev]{schuhmann2022laion}
Christoph Schuhmann, Romain Beaumont, Richard Vencu, Cade Gordon, Ross
  Wightman, Mehdi Cherti, Theo Coombes, Aarush Katta, Clayton Mullis, Mitchell
  Wortsman, Patrick Schramowski, Srivatsa Kundurthy, Katherine Crowson, Ludwig
  Schmidt, Robert Kaczmarczyk, and Jenia Jitsev.
\newblock Laion-5b: an open large-scale dataset for training next generation
  image-text models.
\newblock In \emph{Proceedings of the 36th International Conference on Neural
  Information Processing Systems}, Red Hook, NY, USA, 2022. Curran Associates
  Inc.

\bibitem[Seo et~al.(2023)Seo, Uzunbas, Han, Cao, Zhang, Tian, and
  Lim]{seo2023online}
Seonguk Seo, Mustafa~Gokhan Uzunbas, Bohyung Han, Sara Cao, Joena Zhang,
  Taipeng Tian, and Ser-Nam Lim.
\newblock Online backfilling with no regret for large-scale image retrieval.
\newblock \emph{arXiv preprint arXiv:2301.03767}, 4, 2023.

\bibitem[Sharma et~al.(2018)Sharma, Ding, Goodman, and Soricut]{sharma2018cc3m}
Piyush Sharma, Nan Ding, Sebastian Goodman, and Radu Soricut.
\newblock Conceptual captions: A cleaned, hypernymed, image alt-text dataset
  for automatic image captioning.
\newblock In \emph{Proceedings of the 56th Annual Meeting of the Association
  for Computational Linguistics (Volume 1: Long Papers)}, pages 2556--2565,
  2018.

\bibitem[Shen et~al.(2020)Shen, Xiong, Xia, and Soatto]{shen2020bct}
Yantao Shen, Yuanjun Xiong, Wei Xia, and Stefano Soatto.
\newblock Towards backward-compatible representation learning.
\newblock In \emph{2020 IEEE/CVF Conference on Computer Vision and Pattern
  Recognition (CVPR)}, pages 6367--6376. IEEE, 2020.

\bibitem[Tang and Wang(2018)]{tang2018ranking}
Jiaxi Tang and Ke Wang.
\newblock Ranking distillation: Learning compact ranking models with high
  performance for recommender system.
\newblock In \emph{Proceedings of the 24th ACM SIGKDD international conference
  on knowledge discovery \& data mining}, pages 2289--2298, 2018.

\bibitem[Thapliyal et~al.(2022)Thapliyal, Tuset, Chen, and
  Soricut]{thapliyal2022xm3600}
Ashish~V Thapliyal, Jordi~Pont Tuset, Xi Chen, and Radu Soricut.
\newblock Crossmodal-3600: A massively multilingual multimodal evaluation
  dataset.
\newblock In \emph{Proceedings of the 2022 Conference on Empirical Methods in
  Natural Language Processing}, pages 715--729, 2022.

\bibitem[Tschannen et~al.(2025)Tschannen, Gritsenko, Wang, Naeem,
  Alabdulmohsin, Parthasarathy, Evans, Beyer, Xia, Mustafa,
  et~al.]{tschannen2025siglip2}
Michael Tschannen, Alexey Gritsenko, Xiao Wang, Muhammad~Ferjad Naeem, Ibrahim
  Alabdulmohsin, Nikhil Parthasarathy, Talfan Evans, Lucas Beyer, Ye Xia, Basil
  Mustafa, et~al.
\newblock Siglip 2: Multilingual vision-language encoders with improved
  semantic understanding, localization, and dense features.
\newblock \emph{arXiv preprint arXiv:2502.14786}, 2025.

\bibitem[van~den Oord et~al.(2019)van~den Oord, Li, and Vinyals]{oord2018cpc}
Aaron van~den Oord, Yazhe Li, and Oriol Vinyals.
\newblock Representation learning with contrastive predictive coding, 2019.

\bibitem[Vejendla(2025)]{vejendla2025drift}
Harshil Vejendla.
\newblock Drift-adapter: A practical approach to near zero-downtime embedding
  model upgrades in vector databases.
\newblock In \emph{Proceedings of the 2025 Conference on Empirical Methods in
  Natural Language Processing}, pages 15949--15960, 2025.

\bibitem[Wu et~al.(2023)Wu, Peng, Zhou, Xiao, Liu, Yuan, Xuan, Valenzuela,
  Chen, Wang, et~al.]{wu2023tinyclip}
Kan Wu, Houwen Peng, Zhenghong Zhou, Bin Xiao, Mengchen Liu, Lu Yuan, Hong
  Xuan, Michael Valenzuela, Xi~Stephen Chen, Xinggang Wang, et~al.
\newblock Tinyclip: Clip distillation via affinity mimicking and weight
  inheritance.
\newblock In \emph{2023 IEEE/CVF International Conference on Computer Vision
  (ICCV)}, pages 21913--21923. IEEE, 2023.

\bibitem[Xu et~al.(2024)Xu, Xie, Tan, Huang, Howes, Sharma, Li, Ghosh,
  Zettlemoyer, and Feichtenhofer]{xu2024metaclip}
Hu Xu, Saining Xie, Xiaoqing Tan, Po-Yao Huang, Russell Howes, Vasu Sharma,
  Shang-Wen Li, Gargi Ghosh, Luke Zettlemoyer, and Christoph Feichtenhofer.
\newblock Demystifying clip data.
\newblock In \emph{International Conference on Learning Representations}, pages
  47812--47831, 2024.

\bibitem[Yang et~al.(2024)Yang, An, Huang, Bi, Yu, Yang, Diao, and
  Xu]{yang2024clipkd}
Chuanguang Yang, Zhulin An, Libo Huang, Junyu Bi, Xinqiang Yu, Han Yang, Boyu
  Diao, and Yongjun Xu.
\newblock Clip-kd: An empirical study of clip model distillation.
\newblock In \emph{2024 IEEE/CVF Conference on Computer Vision and Pattern
  Recognition (CVPR)}, pages 15952--15962. IEEE, 2024.

\bibitem[You et~al.(2017)You, Xu, Xu, and Tao]{you2017multiteacher}
Shan You, Chang Xu, Chao Xu, and Dacheng Tao.
\newblock Learning from multiple teacher networks.
\newblock In \emph{Proceedings of the 23rd ACM SIGKDD international conference
  on knowledge discovery and data mining}, pages 1285--1294, 2017.

\bibitem[Young et~al.(2014)Young, Lai, Hodosh, and
  Hockenmaier]{young2014flickr}
Peter Young, Alice Lai, Micah Hodosh, and Julia Hockenmaier.
\newblock From image descriptions to visual denotations: New similarity metrics
  for semantic inference over event descriptions.
\newblock \emph{Transactions of the association for computational linguistics},
  2:\penalty0 67--78, 2014.

\bibitem[Zhai et~al.(2023)Zhai, Mustafa, Kolesnikov, and Beyer]{zhai2023siglip}
Xiaohua Zhai, Basil Mustafa, Alexander Kolesnikov, and Lucas Beyer.
\newblock Sigmoid loss for language image pre-training.
\newblock In \emph{2023 IEEE/CVF International Conference on Computer Vision
  (ICCV)}, pages 11941--11952, 2023.

\bibitem[Zhang et~al.(2022)Zhang, Ge, Shen, Li, Yuan, Xu, Wang, and
  Shan]{zhang2022hotrefresh}
Binjie Zhang, Yixiao Ge, Yantao Shen, Yu Li, Chun Yuan, Xuyuan Xu, Yexin Wang,
  and Ying Shan.
\newblock Hot-refresh model upgrades with regression-alleviating compatible
  training in image retrieval.
\newblock \emph{arXiv preprint arXiv:2201.09724}, 2022.

\bibitem[Zhang et~al.(2023{\natexlab{a}})Zhang, Ge, Xu, Shan, and
  Shou]{zhang2023taca}
Binjie Zhang, Yixiao Ge, Xuyuan Xu, Ying Shan, and Mike~Zheng Shou.
\newblock Taca: Upgrading your visual foundation model with task-agnostic
  compatible adapter.
\newblock \emph{arXiv preprint arXiv:2306.12642}, 2023{\natexlab{a}}.

\bibitem[Zhang et~al.(2023{\natexlab{b}})Zhang, Su, Ge, Xu, Wang, Yuan, Shou,
  and Shan]{zhang2023darwinian}
Binjie Zhang, Shupeng Su, Yixiao Ge, Xuyuan Xu, Yexin Wang, Chun Yuan,
  Mike~Zheng Shou, and Ying Shan.
\newblock Darwinian model upgrades: Model evolving with selective
  compatibility.
\newblock In \emph{Proceedings of the AAAI Conference on Artificial
  Intelligence}, pages 3393--3400, 2023{\natexlab{b}}.

\bibitem[Zhou et~al.(2023)Zhou, Li, Shrivastava, Zhao, Torralba, Tian, and
  Lim]{zhou2023bt2}
Yifei Zhou, Zilu Li, Abhinav Shrivastava, Hengshuang Zhao, Antonio Torralba,
  Taipeng Tian, and Ser-Nam Lim.
\newblock Btˆ 2: Backward-compatible training with basis transformation.
\newblock In \emph{2023 IEEE/CVF International Conference on Computer Vision
  (ICCV)}, pages 11195--11204. IEEE, 2023.

\end{thebibliography}
